\documentclass[11pt]{article}
\newcommand{\nametag}[1]{#1}

\usepackage[preprint]{acl}

\usepackage{times}
\usepackage{latexsym}

\usepackage[T1]{fontenc}
\usepackage[utf8]{inputenc}

\usepackage{microtype}

\usepackage{inconsolata}

\usepackage{graphicx}

\usepackage[inline]{enumitem}
\usepackage{microtype}
\usepackage{graphicx}
\usepackage{booktabs}
\usepackage{amsmath}
\usepackage{amssymb}
\usepackage{multirow}
\usepackage{pifont}
\usepackage{tikz}
\usetikzlibrary{calc,positioning,arrows.meta,fit,backgrounds,shapes.geometric}
\usepackage{tabularx}
\usepackage{arydshln} 
\usepackage{subcaption}
\usepackage{float}
\newcommand{\correspondingauthor}[1]{\nametag{\thanks{#1}}}

\usepackage{makecell}
\newcommand{\deltacell}[2]{\textit{#1 {\scriptsize(#2)}}}
\title{Universal Boosts, Specific Suppressors: Sparse Autoencoder Steering of Medical Vision-Language Models}

\author{
\textbf{Farhad Nooralahzadeh\textsuperscript{1,6}\correspondingauthor{Corresponding Author}}\quad \textbf{Benjamin Gundersen\textsuperscript{1}}\quad\textbf{Nicolas Deperrois\textsuperscript{1}}\\
\textbf{Hidetoshi Matsuo\textsuperscript{2}}\quad \textbf{Mizuho Nishio\textsuperscript{2}}
\quad \textbf{Thomas Frauenfelder\textsuperscript{1}}\quad \textbf{Ahmed Allam\textsuperscript{1}}\\
\textbf{Christian Bl\"uthgen\textsuperscript{5}}\quad \textbf{Michael Moor\textsuperscript{3,4}} \quad \textbf{Michael Krauthammer\textsuperscript{1}} \\
\textsuperscript{1}University of Zurich and University Hospital of Zurich, Switzerland\\
  \textsuperscript{2}Kobe University, Japan \quad\textsuperscript{3}ETH AI Center, Switzerland  \quad\textsuperscript{4}ETH Zurich, Switzerland\\
  \textsuperscript{5}Stanford University, USA \quad\textsuperscript{6}Zurich University of Applied Sciences, Switzerland\\
 }
\begin{document}
\maketitle
\begin{abstract}

% \begin{itemize}
% \item  SAEs on the language residual stream give sparse directions you can edit at decode time; no weight updates.
% \item Boost features are functionally cross-architecture; suppress features are model-specific (quantify: Jaccard + cosine band from census).
% \item On MIMIC-CXR test , per-model SAE steering improves Composite on RadVLM, LLaVA-Rad, CheXOne; all five metrics can move together on at least one config per model; bootstrap p on GREEN for each.

% \item This asymmetry explains why naive transfer / cross-model suppress lists fail while per-model pipeline works; motivates decomposed steering (universal boost + per-model suppress).

% \item Cost order-of-magnitude vs GRPO;
% \item code (if the polished one be available)
% \end{itemize}
Medical vision-language models (VLMs) often hallucinate findings when generating chest X-ray reports: they fabricate findings that are not present in the image, miss important ones, or locate them incorrectly. We mitigate this \emph{without} weight updates by decoding-time residual steering on a per-token sparse autoencoder (SAE) basis: Top-$K$ SAEs on late layers, causal steering against clinical errors, then combined \emph{suppress/boost} intervention at inference time. 
On the MIMIC-CXR test split, our inference-only method improves the quality of generated reports for three radiology VLMs (RadVLM, LLaVA-Rad, and CheXOne), with relative improvements of +5.4\%, +7.2\%, and +17.0\% in the clinical composite metric, and statistically significant GREEN gains on all backbones. A cross-model feature alignment shows that the quality-promoting~(\emph{boost}) directions overlap strongly across architectures, whereas hallucination-linked~(\emph{suppress}) directions are model-specific. Therefore, transferable steering must treat suppression per-backbone, rather than sharing a universal suppress list. 
The same recipe transfers zero-shot to IU-Xray (Green $+7.7\%$~rel.) without retraining, confirming that the identified features are properties of the model, not of the training corpus.
We release causal feature sets and an interactive feature dashboard~\footnote{\url{https://cxr-sparse-feature-dashboard.netlify.app/}}.
\end{abstract}

\section{Introduction}
Medical vision-language models (VLMs) can hallucinate findings, omit pathologies, or mislocalize observations during the generation of radiology reports from medical images \citep{calamida2023radiology,ostmeier-etal-2024-green}. The most common approach to mitigate this problem is post training-time optimization using Reinforcement Learning (RL) by employing Direct Preference Optimization~(DPO) or Group Relative Policy Optimization~(GRPO) fine-tuning on medical VLMS, such as  CheXalign~\citep{hein-etal-2025-chexalign}, RadVLM-GRPO \citep{gundersen2026radvlmgrpo}, and CheXOne \citep{zhang2026reasoningenabledvisionlanguagefoundationmodel}, which achieved strong gains. However, these remedies require gradient updates, optimized reward signals, and per-model training runs, which result in substantial computational overhead (e.g., $\sim$350 H200 GPU-hours for RadVLM-GRPO; \citealp{gundersen2026radvlmgrpo}). 
% \begin{figure}
%     \centering
%     \includesvg[width=1\linewidth]{figures/SAE-RadVLM-fig1_teaser}
%    \caption{\textbf{SAE hallucination mitigation on a real CXR.}
% The unsteered RadVLM hallucinates
% {\tiny\sffamily\colorbox{yellow!25}{WRONG SEVERITY}} and
% {\tiny\sffamily\colorbox{red!12}{FALSE FINDING}}.
% SAE steering zeroes harmful and amplifies beneficial features
% ($\boldsymbol{\delta}$ added to hidden state; weights frozen).
% The steered report matches ground truth. MIMIC-CXR test set.
% }
%     \label{fig:sae-radvlm}
% \end{figure}

\begin{figure}
    \centering
\includegraphics[width=1\columnwidth]{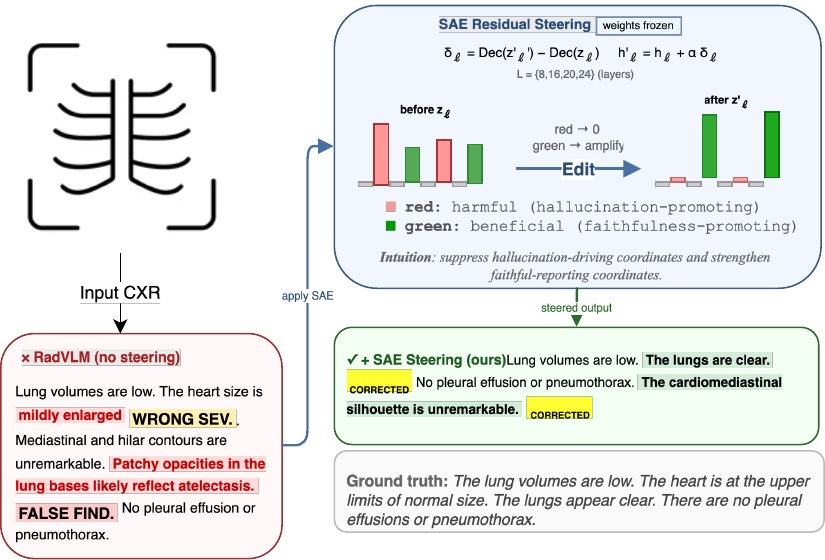}
    \caption{%
\textbf{SAE hallucination mitigation on a real CXR (Schematic illustration).}
The unsteered RadVLM hallucinates
{\tiny\sffamily\colorbox{yellow!25}{WRONG SEVERITY}} and
{\tiny\sffamily\colorbox{red!12}{FALSE FINDING}}.
Each SAE feature is classified as \textcolor{red!70!black}{harmful} or \textcolor{green!50!black}{beneficial} by a \emph{prior causal screen}: zeroing the feature on a validation set and measuring whether clinical errors (GREEN) increase or decrease (\S\ref{method:features}).
In inference, harmful features are zeroed, beneficial ones amplified, and the resulting residual $\boldsymbol{\delta}$ is added to the frozen hidden state.
The steered report matches ground truth.
}
\label{fig:sae-radvlm}
\end{figure}
\begin{figure*}
    \centering
\includegraphics[width=.9\textwidth]{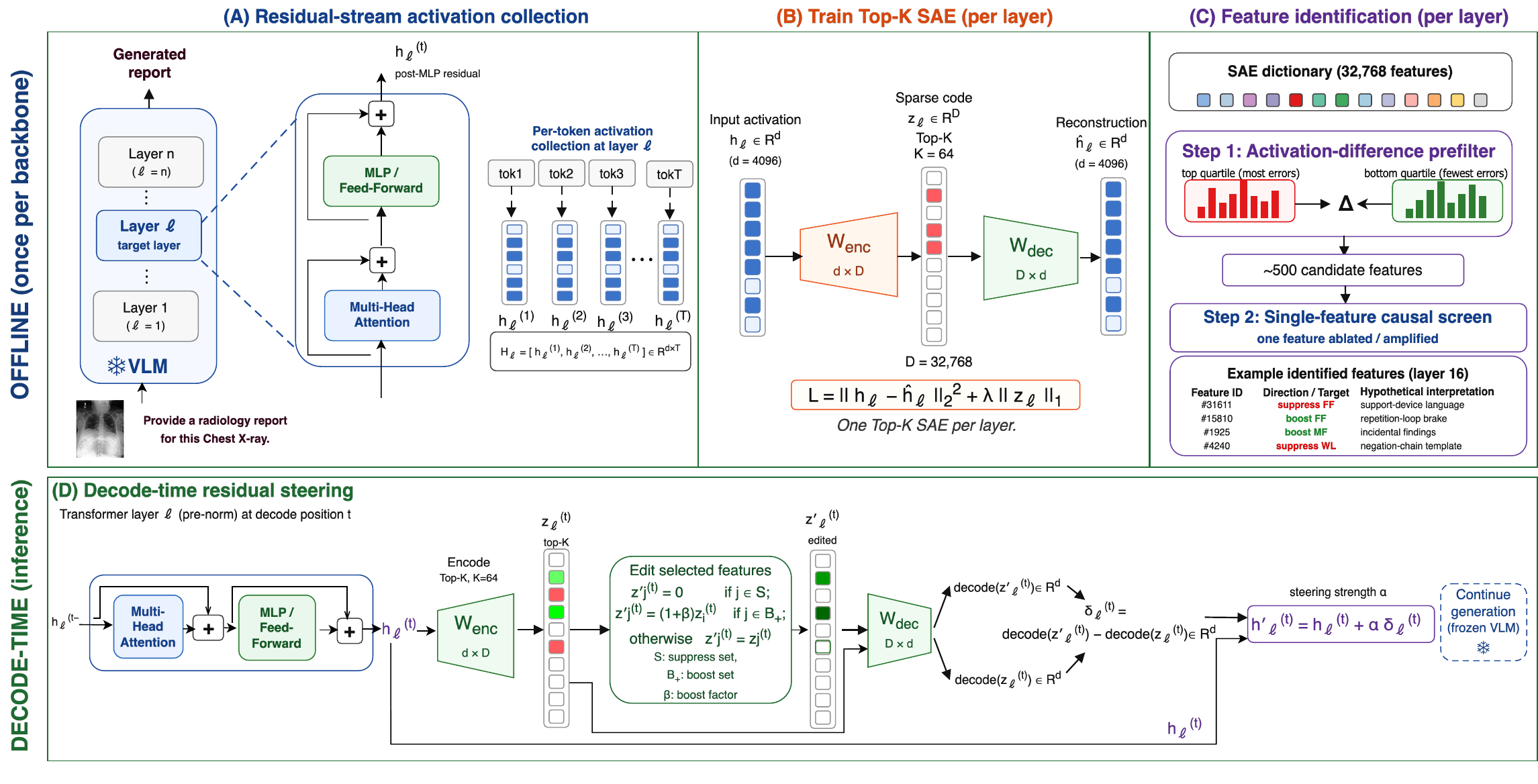}
    \caption{End-to-end SAE residual-steering pipeline: (weights frozen).\textbf{(A)}~Collect per-token residual activations $\mathbf{h}_\ell^{(t)}$ while the VLM generates reports.
\textbf{(B)}~Train a layer-wise Top-$K$ SAE to obtain sparse codes $\mathbf{z}_\ell$.\textbf{(C)}~Identify which coordinates to edit.
Validation studies are ranked by GREEN error count and split into quartiles. \emph{Step~1:} keep ${\sim}500$ features with the largest $|\bar{z}_j^{\text{high}}-\bar{z}_j^{\text{low}}|$, i.e.\ the mean-activation gap between the top quartile (most errors) and bottom quartile (fewest errors) \emph{Step~2:} for each candidate, ablate it on held-out decodes and measure the change in per-type GREEN errors; ablation that \emph{lowers} errors $\Rightarrow$ \textcolor{red!70!black}{suppress}, ablation that \emph{raises} errors $\Rightarrow$ \textcolor{green!50!black}{boost}.
\textbf{(D)}~At inference, apply those edits to $\mathbf{z}_\ell^{(t)}$ (zero suppress, scale boost), decode, and add only the residual difference $\boldsymbol{\delta}_\ell$ to $\mathbf{h}_\ell^{(t)}$ with strength $\alpha$.
}
    \label{fig:pipline}
\end{figure*}
Here, we take an orthogonal approach, inspired by Sparse Autoencoders (SAEs)~\citep{bricken2023towards}, by freezing the model and \emph{editing the model's internal activations at inference time}. SAEs decompose transformer hidden states into high-dimensional sparse representations \citep{bricken2023towards, huben2024sparse,gao2025scaling} and then produce interpretable features that could be linked to specific model behaviors. Recent work shows that SAE features are also useful for steering when the proper features are selected \citep{arad-etal-2025-saes,o2024steering}.
We apply these findings to the safety-critical domain of medical radiograph report generation, where the interpretability of the machine learning model is important.
We organize this work around four research questions that examine the presence (Q1), causal relevance (Q2), generalizability (Q3), and structure of clinically meaningful SAE features (Q4) in medical radiograph VLMs as follows:

\noindent\textbf{Q1:} Do per-token Top-$K$ SAEs trained on medical VLMs residual streams uncover a small, sparse set of features whose activation has a clinically meaningful effect on \emph{each} clinical error type (false findings, missing findings, wrong location, wrong severity)?

\noindent\textbf{Q2:} Can such features be used to mitigate hallucinations at \emph{inference time only}, by editing the residual stream in the SAE basis, without any weight update, or change to the prompt or the decoding algorithm?

\noindent\textbf{Q3:} Does the same identification and steering recipe transfer across three medical VLM backbones that differ in size, vision encoder, and training regime, and does it remain effective \emph{on top of} a reinforcement-learning fine-tune?  Does it also transfer \emph{across datasets}, from MIMIC-CXR (where features are identified) to the independently collected IU-Xray, with no re-training?

\noindent\textbf{Q4:} Looking across these three architectures, do they share the same hallucination-mitigating mechanisms, or do quality-promoting and hallucination-linked directions decompose differently?

In \S\ref{method}, we answer Q1 and Q2 mechanically: per-token Top-$K$ SAEs, an activation-difference prefilter, single-feature causal screening per error type, and combined suppress/boost residual edits at layers: $\{8,16,20,24\}$.
In \S\ref{results}, we answer Q2 and Q3 on the MIMIC-CXR test split, with $(\alpha,K)$ frozen on the validation set (\S\ref{sec:res_generalization}, \S\ref{Per-error-type}). Furthermore, zero-shot cross-dataset transfer to IU-Xray (see Appendix~\ref{sec:appendix_iuxray}). \S\ref{sec:res_census} answers Q4 via a cross-model functional overlap analysis and a per-token decoding-position study.

\section{Related work}
\textbf{Sparse autoencoders for interpretability.} Dictionary learning over transformer activations recovers sparse, interpretable features
\citep{bricken2023towards,huben2024sparse, gao2025scaling,templeton2024scaling}.
\citet{marks2025sparse} by utilizing single-feature interventions to trace causal circuits at the level of individual SAE units and exploring which circuit a feature participates in. \\
We adopt the single-feature intervention approach to answer a different question: \emph{which clinical error type does an SAE feature causally control?} \\
\textbf{Inference-time activation editing.} 
Activation addition~\citep{postmus2024steering}, representation engineering~\citep{zou2023transparency}, and inference-time intervention (ITI)~\citep{li2023inferencetime} all edit hidden states without changing weights.
SAE-based steering refines this line of work by operating in a sparse, interpretable basis: \citet{arad-etal-2025-saes} shows that the choice of feature is decisive, and \citet{o2024steering} applies SAE steering to control refusal in language models.\\
We extend the SAE-steering line into a structured clinical setting with multiple error types and multi-metric evaluation, making the cross-architecture census the centerpiece rather than a side observation.\\
\textbf{SAE steering of vision-language models.}
\citet{pach2026sparse} train SAEs on the \emph{vision encoder} of CLIP and erase visual concepts (e.g., \textit{knives}, \textit{laptops}) in LLaVA's generated descriptions by single-neuron suppression.
Their intervention comes before ours in the process; we are the language-stream counterpart in a clinical setting, residual edits on the LLM, structured clinical error types, and feature selection by causal screening against GREEN~\citep{ostmeier-etal-2024-green} rather than visual monosemanticity.\\
\textbf{Hallucination in medical VLMs.}
Most remedies are applied at training-time such as (a) supervised fine-tuning on curated reports
(LLaVA-Rad \citet{ZambranoChaves2025}; RadVLM~\citep{deperrois2025radvlm}; MAIRA-2~\citep{Bannur2024-ek}), (b) clinical RL with reward engineering (CheXOne~\citep{zhang2026reasoningenabledvisionlanguagefoundationmodel};
RadVLM-GRPO~\citep{gundersen2026radvlmgrpo} ), or (c) region-guided grounding (RGRG~\citep{tanida2023interactive}); these approaches require gradient updates and model-specific rewards, grounding, or fine-tuning signals.
Other alternatives such as decoding-time interventions, reshape next-token logits rather than hidden states (VCD~\cite{leng2024mitigating}; OPERA~\citep{huang2024opera}; DoLa~\citep{chuang2024dola}), but are validated on general-domain benchmarks and do not separate clinical error types. We instead edit the residual stream on an SAE basis: an inference-only method that exposes interpretable per-GREEN-error-type controls, stacks on medical VLMs, and is evaluated with structured clinical metrics.\\
To our knowledge, this is the \textbf{first application of SAE-based steering to medical report generation}, and we further show that a per-model SAE pipeline transfers cleanly onto a GRPO-tuned medical VLM (CheXOne).
\section{Method} \label{method}
Our approach intervenes on a frozen radiology VLM at decoding time and consists of four stages as shown in Figure~\ref{fig:pipline}: 

\paragraph{A) Residual-stream Activation Collection}
Let $\mathbf{h}_\ell^{(t)}$ denote the hidden state at transformer layer $\ell$ and token position $t$ during report generation.
We save per-token activation for all layers $\mathcal{L}$, which is used to \emph{train} sparse autoencoders (SAEs), and a set of layers for intervention hooks, across training and validation-split samples. These tensors correspond to the SAE training datasets and the intervention locations.
\paragraph{(B) Sparse Autoencoder Training}
At each layer $\ell$, a Top-$K$ SAE encodes $\mathbf{h}_\ell^t$ into a sparse representation $\mathbf{z}_\ell^{(t)} \in \mathbb{R}^D$ with at most $K$ nonzeros and decodes $\hat{\mathbf{h}}_\ell^{(t)} = \mathrm{decode}_\ell(\mathbf{z}_\ell^{(t)})$~\citep{gao2025scaling}.
We use dictionary size $D{=}32,768$, sparsity $K{=}64$, and model hidden dimension $d{=}4,096$.
\paragraph{(C) Feature Identification} \label{method:features}
We use two approaches to select features to suppress or amplify. 

\noindent \textit{1. Correlation based selection (Corr) --} 
For each layer $\ell$ and feature $j$, we compute the Pearson correlation between the feature's activation magnitude on the training set and the corresponding GREEN error count (overall or per error type).
Features with the most positive correlation become candidates for \emph{suppression}; the most negatively correlated become \emph{boost} candidates.
The procedure yields per-error-type lists out of the box.

\noindent \textit{2. Causal selection (Causal) --}
This involves a two-stage procedure. First, an \emph{activation-difference prefilter}: we keep the top ${\sim}500$ features exhibiting the largest absolute gap in mean activation between top-quartile (high-error) and bottom-quartile (low-error) samples.
Second, a \emph{single-feature causal screen}: for each candidate, we run a one-feature intervention (zero or amplify) on a held-out validation set and measure the change in the automatic clinical metrics directly~\citep{arad-etal-2025-saes}.
For each surviving feature $j$, we record the \emph{per-error-type causal data} as:
\begin{equation}\small
  \Delta_t^{(j)} = \frac{1}{N}\sum_{i=1}^{N}\big(\#\text{GREEN}_t(a_i^{(j)}) - \#\text{GREEN}_t(b_i)\big)
\end{equation}
for $t\in\{\mathrm{FF},\mathrm{MF},\mathrm{WL},\mathrm{WS}\}$, where $b_i$ and $a_i^{(j)}$ are the baseline and feature-$j$-ablated decodes of validation study $i$, and $\#\text{GREEN}_t(\cdot)$ is the count of GREEN clinically-significant errors: \emph{false findings} (FF), \emph{missing findings} (MF), \emph{wrong location} (WL), and \emph{wrong severity} (WS) (see \S~\ref{metrics}).
$\Delta_t^{(j)}$ represents the mean errors per report. Features are then ranked by their measured causal effect on each error type, not by association alone.

Both methods produce, per layer, two ordered lists including features to \textbf{suppress} ($\Delta_t^{(j)}<0$ associated with errors) and features to \textbf{boost} ($\Delta_t^{(j)}>0$ associated with quality), organized across the four GREEN error types.

\paragraph{(D) Residual Steering}
Given our identified feature lists, we apply inference-time steering to the model's residual stream. We emphasize a \emph{residual} update rather than standard activation patching to prevent SAE reconstruction error from degrading generation quality. 

\noindent \textit{1. Feature Aggregation \& Conflict Resolution --}
At inference time, we truncate our ranked lists to a validation-chosen budget $K$. For causal selection, we aggregate the top-$K$ features across all GREEN error types into a master suppress list ($\mathcal{S}_\ell$) and a master boost list ($\mathcal{B}_\ell$). Because a feature may inadvertently appear on both lists due to differing error profiles, we resolve conflicts by strictly prioritizing suppression to minimize clinical hallucinations: any overlapping feature is removed from $\mathcal{B}_\ell$.

\noindent \textit{2. Feature Editing --}
Let $\mathbf{z}_\ell$ denote the SAE representation of the hidden state $\mathbf{h}_\ell$. We construct the edited features $\mathbf{z}'_\ell$ by forcefully zeroing out suppressed features and scaling boosted features by a factor of $(1+\beta)$:
\begin{equation}\small
  \label{eq:edit}
z'_{\ell, j} =
\begin{cases}
0 & \text{if } j \in \mathcal{S}_\ell \\
(1+\beta)\, z_{\ell, j} & \text{if } j \in \mathcal{B}_\ell \\
z_{\ell, j} & \text{otherwise.}
\end{cases}
\end{equation}
In our experiments, $\beta = 1.0$, doubling boosted coordinates, while baseline unsteered sets use $\alpha = 0$ and $\beta = 0$).

\noindent \textit{3. Applying the Residual Update --}
Standard SAE-mediated activation patching~\citep{marks2025sparse,templeton2024scaling} replaces the original hidden state with the SAE reconstruction:
\begin{equation}
  \label{eq:sae_patch}\small
  \mathbf{h}'_\ell \;=\; (1{-}\alpha)\,\mathbf{h}_\ell + \alpha\,\mathrm{decode}_\ell(\mathbf{z}'_\ell)
\end{equation}
However, this injects baseline reconstruction error ($\mathrm{decode}_\ell(\mathbf{z}_\ell){-}\mathbf{h}_\ell$) directly into the network. 

To avoid this, we apply a pure residual update. We isolate the exact delta caused by our edits and add it directly back to the original hidden state, scaled by steering strength $\alpha$:
\begin{equation}\small
  \label{eq:residual}
  \boldsymbol{\delta}_\ell = \mathrm{decode}_\ell(\mathbf{z}'_\ell) - \mathrm{decode}_\ell(\mathbf{z}_\ell), \quad
  \mathbf{h}'_\ell = \mathbf{h}_\ell + \alpha \, \boldsymbol{\delta}_\ell
\end{equation}
This formulation is the SAE-basis analogue of additive inference-time edits~\citep{li2023inferencetime}: it exactly cancels out the reconstruction residue and preserves the unedited knowledge in $\mathbf{h}_\ell$. We retain Eq.~\eqref{eq:sae_patch} only as an early ablation (it degrades sharply for $\alpha\!\geq\!0.25$ in our setting) and apply Eq.~\eqref{eq:residual} at every layer in $\mathcal{L}$ for every generated token.

\section{Experiments}
\paragraph{Dataset and Metrics.} \label{metrics}
We use \textbf{MIMIC-CXR-JPG}~\citep{PhysioNet-mimic-cxr-2.1.0}, a large public chest-radiograph dataset paired with radiology reports.
Following \citet{deperrois2025radvlm}, we restrict to single frontal chest X-rays paired with findings,
where references to prior studies are removed.
After filtering there are 230,980 image-text pairs in the training/validation set and 3,314 in the test set. The corresponding datapoints can be extracted from the RadVLM instruction dataset \citep{deperrois2025radvlminstructiondataset}.
SAE training and feature identification use the training split; hyperparameter tuning ($\alpha$, number of features, steering mode) uses the validation split; test-set numbers are reported once, on a frozen configuration.

To test whether features identified on MIMIC-CXR transfer to an unseen corpus, and to provide DUA-compliant qualitative examples (MIMIC-CXR's data-use agreement prohibits reproducing images and reports in publications), we additionally evaluate on \textbf{IU-Xray}~\citep{demner2016preparing} ($N=3,307$ frontal studies after filtering for image availability).
No SAE retraining, feature re-screening, or hyperparameter re-tuning is performed: the same weights, feature lists, and $(\alpha,K)$ from MIMIC-CXR are applied zero-shot.
Results are reported in Appendix~\ref{sec:appendix_iuxray}.

We primarily employ GREEN \citep[][based on \emph{StanfordAIMI/GREEN-radllama2-7b}]{ostmeier-etal-2024-green}, which classified errors into six categories: \emph{False Findings} (FF): False report of a finding in the candidate, \emph{Missing Findings} (MF): Missing a finding present in the reference, \emph{Wrong Location} (WL): Misidentification of a finding's anatomic location/position, \emph{Wrong Severity} (WS): Misassessment of the severity of a finding, \emph{False Comparisons} (FC): Mentioning a comparison that isn't in the reference, and \emph{Missing Comparisons} (MC): Omitting a comparison detailing a change from a prior study. 

We focus on FF, MF, WL, and WS since references to prior studies were filtered. We report the GREEN score.
Further, we report other clinical metrics \texttt{RadGraph-F1 simple}~\citep{jain2021radgraph}, \texttt{CheXbert micro F1, 14 labels} ~\citep{smit-etal-2020-combining}, and \texttt{BERTScore}~\citep{Zhang2020BERTScore}. We formulate the \texttt{Composite Score} as $ 0.4 \times GREEN+0.3\times RadGraph+0.2\times CheXbert+0.1  \times BERTScore$. We also report \texttt{RadCliQ} \cite{Yu2022.08.30.22279318}, a learned index that combines RadGraph, BERTScore, CheXbert embeddings, and BLEU, calibrated to radiologist preferences. Furthermore, \texttt{BLEU-4} and  \texttt{ROUGE-L} are evaluated as general text generation metrics.
These scores are calculated using the RadEval framework \citep{xu-etal-2025-radeval}.

\paragraph{Base Models.} The primary VLM we use is {RadVLM} (Qwen-3-VL-8B backbone, Radiology instruction-tuned). All weights in the model remain frozen throughout the pipeline.
For generalization aspects, we choose {LLaVa-Rad} (7B, LLaVa/Vincua) and {CheXOne} (Qwen2.5-VL-3B, GRPO-tuned). For each mode, we report unsteered test inference and SAE residual steering using SAEs and targeted features trained on that model's own activations, with hyperparameters chosen on the validation set. CheXOne is evaluated in instruct mode (standard findings prompt, no long reasoning trace) for comparability with the other report generation models.

\subsection{Experimental Pipeline}
We run the end-to-end recipe from \S~\ref{method} as a sequence of separated jobs: activations collection $\rightarrow$ SAE training  $\rightarrow$ feature selection $\rightarrow$ hyperparameter search on validation $\rightarrow$ generate reports on test set. We also perform the same recipe on LLaVA-Rad and CheXOne.

\paragraph{End-to-end Configuration}\label{sec:method_config}
The following setups define the pipeline we use for all main experiments (RadVLM, unless we note otherwise) on top of the methods explained previously. For {SAE training} and per-token activation collection, we use a fixed subset of $3,000$ studies drawn \emph{only} from the {training} split.
Indices are selected with \emph{stratified sampling} over CheXpert-style pathology groups (primary label per study, with a minimum count per group) and a fixed random seed, so rare findings are represented rather than a random slice dominated by common classes. We limit the generated reports (decoding length) to 512 tokens, and we train one Top-$K$ SAE per SAE training layer (e.g, ten layers $\{0,4,8,12,16,20,24,28,32,35\}$  for RadVLM's 36-block stack). We attach steering hooks to the four mid/late layers  $\mathcal{L}{=}\{8,16,20,24\}$, where SAE reconstruction quality is high, and activation norms remain moderate enough for steering perturbations to be effective (Appendix \ref{sec:appendix_layer_selection}, Table \ref{tab:layer_quality}, as layer ablations).

Correlation-based selection uses the same 3,000 examples by later activation files paired with per-sample GREEN error count. 
In the causal selection process, the prefilter runs over all $D$ dictionary features at each of the four hook layers, then the single-feature forward
screens on only $100$ held-out validation examples. We split validation studies by RadCliQ quality into four quartiles (worst 25\%, …, best 25\%), then draw roughly equal counts from each quartile, so that the panel mixes low- and high-quality reports rather than being dominated by one regime.
The target features are recorded as per-layer JSON files (top-$K$ indices for suppress and boost, optionally by GREEN type).
\paragraph{Hyperparameter search.} For each model backbone after activation collection and SAEs training, we perform a grid search over the subset of validation split (200 stratified samples), over steering strength
$\alpha \in \{0.10, 0.15, 0.20, 0.25, 0.30, 0.35, 0.40, 0.50\}$,
and per-layer feature count $K \in \{20,50,100\}$, steering approach (residual vs.\ blend; Combined vs.\ suppress-only vs.\ boost-only),
the layer subset/multi-layer ablation $\mathcal{L}$ ($\{8,16,20,24\}$),
and the boost factor $\beta \in \{0.5, 1.0\}$. Each condition runs a full report generation on all validation cases with SAE hooks at the four hook at the four steered layers and is scored with 
GREEN, RadGraph, CheXbert, and BERTScore.

\begin{table*}[t]
  \centering
  \footnotesize
   \setlength{\tabcolsep}{4pt}
  \renewcommand{\arraystretch}{1.05}
  \resizebox{1\textwidth}{!}{%
  \begin{tabular}{@{}p{5cm}cccccccc@{}}
    \toprule
    \textbf{Setting} & \textbf{GREEN ($GR$}$\uparrow$ & \textbf{RG}$\uparrow$ & \textbf{CXB$_{\mu}$}$\uparrow$ & \textbf{BS}$\uparrow$ & 
    \textbf{BL}$\uparrow$ & \textbf{RL}$\uparrow$ & \textbf{Comp.}$\uparrow$ & \textbf{RCQ}$\uparrow$ \\
    \midrule

    \textbf{RadVLM}                                 & 27.7  & 18.6 & 47.7 & 52.9 & 6.6 & 25.8 & 31.5 & $-$1.193 \\
    \quad + SAE Combined {\scriptsize ($\alpha{=}0.20$, $K{=}20$)}                       & \textbf{28.8}\rlap{$^{\ast}$}  & \textbf{21.0} & \textbf{50.3} & \textbf{53.0}&\textbf{7.0} & 25.8 & \textbf{33.2} & \textbf{$-$1.159} \\
    \hdashline
\quad \textit{$\Delta$ pp / abs.\ (rel.\%)} &
\deltacell{$+1.0$}{$+3.9\%$} &
\deltacell{$+2.4$}{$+12.9\%$} &
\deltacell{$+2.6$}{$+5.5\%$} &
\deltacell{$+0.1$}{$+0.2\%$} &
\deltacell{$+0.4$}{$+6.1\%$} &
\deltacell{$+0.0$}{$+0.2\%$} &
\deltacell{$+1.7$}{$+5.4\%$} &
\deltacell{$+0.034$}{---} \\
    % \quad \textit{$\Delta$ (pp / abs.)}         & \textit{$+1.0$} & \textit{$+2.4$} & \textit{$+2.6$} & \textit{$+0.1$} & \textit{$+0.4$}& \textit{$+0.0$}& \textit{$+1.7$} & \textit{$+0.034$} \\
    % \quad \textit{rel.\ \%}                     & \textit{$+3.9\%$}  & \textit{$+12.9\%$} & \textit{$+5.5\%$} & \textit{$+0.2\%$} &
    % \textit{$+6.1\%$}&
    % \textit{$+0.2\%$}&
    % \textit{$+5.4\%$} & --- \\
    \midrule

    \textbf{LLaVA-Rad}                       & 28.5  & 16.8 & 53.3 & 48.9 & 5.2 & 22.2 & 33.3 & $-$1.374 \\
    \quad + SAE Combined {\scriptsize ($\alpha{=}0.50$, $K{=}10$)}                                 & \textbf{31.4}\rlap{$^{\ast}$}& \textbf{19.4} & 52.6 & \textbf{54.0} & \textbf{6.3}& \textbf{24.5}& \textbf{35.7} & \textbf{$-$1.276} \\
    \hdashline
\quad \textit{$\Delta$ pp / abs.\ (rel.\%)}&
\deltacell{$+2.9$}{$+10.2\%$} &
\deltacell{$+2.6$}{$+15.4\%$} &
\deltacell{$-0.7$}{$-1.3\%$} &
\deltacell{$+5.1$}{$+10.5\%$} &
\deltacell{$+1.1$}{$+20.8\%$} &
\deltacell{$+2.3$}{$+10.3\%$} &
\deltacell{$+2.4$}{$+7.2\%$} &
\deltacell{$+0.098$}{---} \\
    % \quad \textit{$\Delta$ (pp / abs.)}         & \textit{$+2.9$}  & \textit{$+2.6$} & \textit{$-0.7$} & \textit{$+5.1$} & 
    % \textit{$+1.1$}&
    % \textit{$+2.3$}&
    % \textit{$+2.4$} & \textit{$+0.098$} \\
    % \quad \textit{rel.\ \%}                     & \textit{$+10.2\%$}  & \textit{$+15.4\%$} & \textit{$-1.3\%$} & \textit{$+10.5\%$} & 
    % \textit{$+20.8\%$}&
    % \textit{$+10.3\%$}&
    % \textit{$+7.2\%$} & --- \\
    \midrule

    \textbf{CheXOne}                                  & 22.3  & 18.9 & 35.5 & 49.5 &  4.6& 22.7& 25.0 & $-$1.339 \\
    \quad + SAE Combined {\scriptsize ({$\alpha{=}0.50$, $K{=}100$)}}                        & \textbf{24.9}\rlap{$^{\ast}$}  & \textbf{20.2} & \textbf{38.7} & \textbf{54.6}& \textbf{5.1}&\textbf{23.6}  & \textbf{29.2} & \textbf{$-$1.237} \\
    \hdashline
\quad \textit{$\Delta$ (pp / rel.)} &
\deltacell{$+2.6$}{$+11.7\%$} &
\deltacell{$+1.3$}{$+6.8\%$} &
\deltacell{$+3.2$}{$+9.1\%$} &
\deltacell{$+5.1$}{$+10.3\%$} &
\deltacell{$+0.5$}{$+10.3\%$} &
\deltacell{$+0.9$}{$+4.2\%$} &
\deltacell{$+4.2$}{$+17.0\%$} &
\deltacell{$+0.102$}{---} \\
   %  \hdashline
   %  \quad \textit{$\Delta$ (pp / abs.)}         & \textit{$+2.6$}  & \textit{$+1.3$} & \textit{$+3.2$} & \textit{$+5.1$} &
   %  \textit{$+0.5$}&
   % \textit{$+0.9$} &\textit{$+4.2$} & \textit{$+0.102$} \\
   %  \quad \textit{rel.\ \%}                     & \textit{$+11.7\%$}& \textit{$+6.8\%$} & \textit{$+9.1\%$} & 
   %  \textit{$+10.3\%$} &
   %  \textit{$+10.3\%$}&
   % \textit{$+4.2\%$} &
   %  \textit{$+17.0\%$} & --- \\
    \bottomrule
  \end{tabular}
  }
  \caption{Steered models evaluation results on the MIMIC-CXR test set.  $\Delta$ rows show absolute change in percentage points (pp), It uses full--percision RadEval means before rounding. The $rel.\%$ is relative gain w.r.t.\ the baseline ({\small (SAE-Base)/Base $\times$ 100}). 
  \textbf{RG}: RadGraph F1-simple, 
  \textbf{CXB}$_{\mu}$: CheXbert-14 micro F1, 
  \textbf{BS}: BERTScore F1, 
  \textbf{BL} and \textbf{RL}: BLEU-4 and ROUGE-L,
  \textbf{Comp.}: Composite,
 \textbf{RCQ:} RadCliQ-v1 quality on its native learned scale (higher is better). 
  $\ast$: $p{<}0.001$ paired bootstrap on GREEN (\#resample=10,000).}
  \label{tab:main_results}
\end{table*}
\section {Main results}\label{results}
We organize our empirical findings to discuss the design choices incrementally. We first identify the SAE steering operating point per-model by ablating one design factor at a time on the validation split (\S~\ref{grid_search}), then evaluate the chosen configuration on the held-out test set with paired bootstrap significance (\S~\ref{sec:results}). We then ask \emph{where} the gain comes from by decomposing the test-set delta across GREEN error types and tracing that decomposition down to a single SAE feature (\S~\ref{Per-error-type}). Section~\ref{sec:res_generalization} provides the same pipeline results over two further medical VLMs (LLaVA-RAD and the GRPO-tuned CheXOne). In section~\ref{sec:res_census}, the resulting boost-suppress feature lists across architectures and provide the structure finding that motivates this paper.
We provide qualitative analysis at \S~\ref{sec:res_qualitative}.
In Appendix~\ref{sec:appendix_iuxray}, we report cross-dataset transfer to IU-Xray as dataset generalization.

\subsection{Finding the per-model operating point}
\label{grid_search}
We choose causal selection over a Pearson-correlation proxy because it tells us, for each feature, the effect on each GREEN error type separately ($\Delta_{\mathrm{FF}}, \Delta_{\mathrm{MF}}, \Delta_{\mathrm{WL}}, \Delta_{\mathrm{WS}}$), rather than just an overall association with hallucination.\\
\textbf{Grid Search.} We keep the causal feature ranking and vary and identify the best SAE steering approach on models by varying one design factor at a time on a stratified subset of the validation split ($n=200$) scored with GREEN, RadGraph, CheXbert-14, and BERTScore. We find that for the steering mode, \emph{residual steering} outperforms its counterpart (SAE-mediated activation patch). Furthermore, the \emph{Combined} approach with $\beta=1$ gains on every metric on every backbone; suppress-only and boost-only each regress at least one metric on at least one backbone.
 Among all the $\alpha$ and $K$ combinations, the optimal points are for RadVLM~($\alpha=0.2$, $K=20$), for LLaVA-Rad~($\alpha=0.5$, $K=10$) and for CheXOne~($\alpha=0.5$, $K=100$). We also observe that the optimal $\alpha$ is small and well-defined ($0.20$--$0.50$); increasing it breaks coherent reporting and degrades the models' output distribution. On the same subset, steering at four layers $\{8,16,20,24\}$ beats a single layer and reduced subsets.

\subsection{RadVLM Results}
\label{sec:results}
We evaluate on the held-out MIMIC-CXR test split once with the validation-frozen configuration; Table~\ref{tab:main_results} reports the metrics of that evaluation.

SAE steering improves \emph{every} clinical and text-quality metric on RadVLM (Table \ref{tab:main_results}, top block):
GREEN $27.7 \to 28.8\%$ ($+1.0$ pp absolute, $+3.9\%$ rel.),
RadGraph $18.6 \to 21.0\%$ ($+12.9\%$ rel.),
CheXbert $47.7 \to 50.3\%$ ($+5.5\%$ rel.),
BERTScore $52.9 \to 53.0\%$,
RadCliQ $-1.193 \to -1.159$.
The Composit gains $+1.7$ pp absolute, $+5.4\%$ relative.
NLG metrics also nudge up (BLEU-4 $6.6\!\to\!7.0$; ROUGE-L a bit more than $25.8$). We perform a statistical significance test with paired bootstrap over per-sample scores, resampling $10,000$  from the test set. The GREEN delta is significant at $p<0.001$.
% We do not bootstrap RadGraph, CheXbert, BERTScore, or RadCliQ metrics individually, but we observe that all four together with GREEN move in the right direction at our chosen operating point, and we did not see the same joint pattern for the other setting in our validation grid search. However, the Green weakens slightly from the validation set to test ($+5.7\%$ rel.\ $\to$ $+3.9\%$ rel.), and all five metrics still improve together.
\subsection{Where does the gain come from?
Per-error-type decomposition} 
\label{Per-error-type}
The composite gain on RadVLM, tells us that the mean report quality improves; it does not say \emph{which} errors are removed. We first decompose the test-set delta across the six GREEN categories (Table~\ref{tab:per_error}), then into individual SAE features (Table~\ref{tab:feature_pertype_summary_L16}).

\begin{figure}[t]
\centering
\footnotesize

% ---------- Table 1 ----------
\setlength{\tabcolsep}{4pt}
\resizebox{\columnwidth}{!}{%
\begin{tabular}{@{}lrrr@{}}
\toprule
\textbf{GREEN error type} & \textbf{Baseline} & \textbf{+ SAE Combined} & \textbf{$\Delta$} \\
\midrule
False finding (FF) & 5,305 & 6,417 & $+1112$ \\
Missing finding (MF) & 7,888 & 7,234 & $-654$ \\
Wrong location (WL) & 1,040 & 1,208 & $+168$ \\
Wrong severity (WS) & 881 & 953 & $+72$ \\
False comparison (FC) & 194 & 136 & $-58$ \\
Missing comparison (MC) & 71 & 66 & $-5$ \\
\midrule
\textbf{Total significant errors} & \textbf{15,379} & \textbf{16,014} & \textbf{$+635$} \\
\bottomrule
\end{tabular}%
}

\captionof{table}{Per-error-type breakdown of GREEN \emph{significant} errors for RadVLM on the MIMIC-CXR test set, at the per-token SAE operating point of Table~\ref{tab:main_results}.}
\label{tab:per_error}

\vspace{.5em}

% ---------- Table 2 ----------
\setlength{\tabcolsep}{1.5pt}
\resizebox{\columnwidth}{!}{%
\begin{tabular}{@{}lrrrr@{}}
\toprule
Error type & \#supp. & \#boost & Top supp. & Top boost \\
\midrule
False finding & 100 & 368 & \#31611 & \#15810 \\
Missing finding & 68 & 397  & \#2541 & \#1925\\
Wrong location & 155 & 273  & \#4240 & \#28456 \\
Wrong severity & 450 & 14  & \#29962 & \#7046 \\
\bottomrule
\end{tabular}%
}

\captionof{table}{Per-error-type causal feature decomposition at $\ell=16$ (RadVLM, 500 prefiltered features). The count of features whose single-feature ablation reduced (\emph{\#supp}) or increased (\emph{\#boost}) that error, and the
strongest individual feature in each direction.}
\label{tab:feature_pertype_summary_L16}

\end{figure}

% \noindent\textbf{The improvement on GREEN is an error redistribution, not a free lunch.} Our steering pipeline removes $654$ missing-finding \textbf{(MF)} errors as the most clinically consequential category, together with $58$ false comparision and 5 missing-comparisoon errors.
% At the same time, false-finding \textbf{(FF)} increased by $+1,112$, with smaller rises in wrong location ($+168$) and wrong severity ($+72$). The total significant-error count, therefore, goes up by $+635$.
\paragraph{The gain on GREEN is an error redistribution, not a free lunch.} 
Steering removes $654$ MF errors but adds $+1{,}112$ FF, $+168$ WL, and $+72$ WS errors (Table~\ref{tab:per_error}). GREEN nevertheless improves because it is structurally \emph{asymmetric in MF}: every recovered missing finding simultaneously adds $1$ to the numerator and subtracts $1$ from the denominator, whereas a new FF/WL/WS error only enlarges the denominator.
The net effect is a \textbf{shift toward more complete reporting}: the steered model recovers reference findings that the baseline omits (MF~$\downarrow$, mean GREEN~$\uparrow$), occasionally at the cost of mentioning content the reference does not. RadGraph F1, CheXbert-14, BERTScore, and RadCliQ all move in the same direction (Table~\ref{tab:main_results}), confirming the gain is not metric gaming. We provide an extended analysis of this trade-off in Appendix~\ref{sec:appendix_error_redistribution}.
\paragraph{Single features carry type-specific causal roles.} Table~\ref{tab:feature_pertype_summary_L16} shows that the same SAE coordinate can be a top suppress for one GREEN type and top boost for another (e.g., \#1925 is MF-boost and WL-suppress at $\ell{=}16$; \#30874 is MF-suppress and FF-boost at $\ell{=}20$). Across the 500-feature pool at $\ell{=}16$, 80 coordinates satisfy $\Delta_{\mathrm{FF}}<0$ and $\Delta_{\mathrm{MF}}>0$ simultaneously, justifying separate per-error-type lists in Eq.~\ref{eq:edit}.
The per-layer balance also differs: $\ell{=}24$ is almost entirely FF-boost ($494/500$), while wrong-severity shifts from mixed at $\ell{=}8$ to strongly WS-suppress at $\ell{=}16$ ($450$ vs.\ $14$). This heterogeneity motivates steering at all four layers (detailed feature activation analysis in Appendix~\ref{sec:appendix_error_redistribution}).
\subsection{Cross-architecture generalization}
\label{sec:res_generalization}
We apply the same pipeline to LLaVA-Rad and CheXOne, training one SAE stack and one feature list per backbone and a validation-selected $(\alpha, K)$. Table~\ref{tab:main_results} shows the results.
The composite improves by $+7.2\%$ on LLaVA-Rad and $+17.0\%$ on CheXOne (relative to each model's own baseline).
The largest gain is on CheXOne, which is already GRPO-fine-tuned~\citep{zhang2026reasoning}, so one might expect little room for improvement for an inference-time edit; we observe the opposite. SAE residual steering is therefore complementary to GRPO, not redundant with it.

To verify that the identified features belong to the model rather than artifacts of MIMIC-CXR, we apply the exact SAE weights, feature lists and hyperparameters using RadVLM-steered on IU-Xray. We observe a similar trend as GREEN +3.8 pp (+7.7\% rel.), RadGraph +4.1 pp (+16.7\% rel.), Composite +2.5 pp (+5.4\% rel.). In terms of per-error-type pattern, MF decreased by 850 and total errors dropped by 713. Full metrics, per error breakdown, are in Appendix~\ref{sec:appendix_iuxray}.
\subsection{Cross-Model Feature Alignment: What Generalizes and What Does Not}
\label{sec:res_census}
\begin{table}[t]
  \centering
  \footnotesize
  \setlength{\tabcolsep}{2.5pt}
  \resizebox{\columnwidth}{!}{%
  \begin{tabular}{@{}llcc@{}}
    \toprule
    \textbf{Model pair} & \textbf{Direction} & \textbf{W.Jaccard}$\uparrow$ & \textbf{Cosine}$\uparrow$ \\
    \midrule
    RadVLM vs.\ LLaVA-Rad & Suppress & 0.42\,[0.33,\,0.50] & 0.52\,[0.36,\,0.72] \\
    RadVLM vs.\ CheXOne & Suppress & 0.58\,[0.45,\,0.71] & 0.46\,[0.24,\,0.67] \\
    LLaVA-Rad vs.\ CheXOne & Suppress & 0.41\,[0.25,\,0.62] & 0.25\,[0.12,\,0.40] \\
  \midrule
    RadVLM vs.\ LLaVA-Rad & Boost & 0.65\,[0.60,\,0.75] & 0.93\,[0.91,\,0.96] \\
    RadVLM vs.\ CheXOne & Boost & 0.75\,[0.56,\,0.94] & 0.95\,[0.94,\,0.95] \\
    LLaVA-Rad vs.\ CheXOne & Boost & 0.75\,[0.65,\,0.80] & 0.91\,[0.88,\,0.94] \\
    \bottomrule
  \end{tabular}%
  }
  \caption{Cross-model functional overlap with 95\% bootstrap Confidence intervals (CIs) (mean\,[lo,\,hi]).
  Per-cell aggregate is the mean over steered layers $\{8,16,20,24\}$ of the top-$100$
  features per direction; CIs are percentile-bootstrap intervals from $10,000$ resamples
  of the four layer values with replacement, capturing cross-layer stability. Boost CIs are tight and lie well above suppress CIs in every model pair.}
  \label{tab:census}
\end{table}
Selecting features separately for each model raises a mechanistic question: \emph{how many of the underlying features are actually shared across models?} Since SAE feature indices are not directly comparable across architectures, we shift the analysis to a functional perspective. Concretely, for each pair of models, we compute the weighted Jaccard (Ruzicka) similarity between the GREEN error-type profiles of the top 100 features, stratified by direction and layer (see Appendix \ref{sec:appendix_census_definition}). In other words, we check how similar two models are by looking at their most important features and asking: do they affect the same types of errors? We measure this overlap using weighted Jaccard similarity, based on which error types each feature mainly influences, as measured during feature identification (\S\ref{method:features}).
Across all model pairs, the boost features are quite consistent: about 65–75\% of them overlap, and their overall behavior is very similar (cosine around 0.91–0.95, with 95\% confidence intervals of about $\pm0.03$. It indicates that features that improve quality tend to work in the same way across different models.
In contrast, the suppress features are much less consistent. Their overlap is lower (41–58\% ), and in some cases their behavior differs a lot. For example, between LLaVA-Rad and CheXOne, the similarity drops sharply. This suggests that each model handles hallucinations in its own way, relying on different internal mechanisms.
Importantly, this difference is very clear statistically: for every model pair, the similarity of boost features is always higher than the similarity of suppress features, with no overlap between their confidence ranges. So this pattern is robust, and it’s not just due to picking specific layers.\\
Using the boost–suppress gap to design new steering strategies, such as shared boost priors, error-specific suppression weights, or other census-aligned decompositions, is a natural next step, but we treat it as future work.
\subsection{Qualitative examples}
\label{sec:res_qualitative}  

Due to MIMIC-CXR's data-use agreement, we reproduce RadVLM steering results on IU-Xray. Table ~\ref{tab:iuxray_qual_examples} (Appendix~\ref{sec:appendix_iuxray}) gives four pathological cases with GREEN tagged baseline vs. steered reports. 
% Table~\ref{tab:appendix_qual_examples} (Appendix~\ref{sec:appendix_examples}) reproduces four RadVLM steering cards alongside Figure~\ref{fig:sae-radvlm}: three \emph{improvement} cases where steering recovers missing findings and removes fabricated language ($G$ rises from $+0.17$ to $+0.50$), and one \emph{regression} case where steering shortens an accurate NG-tube description and injects false findings ("mild pulmonary edema", "moderate cardiomegaly"), dropping $G$ from $0.75$ to $0.17$.
% This regression matches the FF$\uparrow$ shift in Table~\ref{tab:per_error}: MF-boost features bias the decoder toward longer, more exhaustive reports, and here the unsteered text was already acceptable, so the extra bias surfaces as new findings without support from the image or the reference.

\section{Discussion and Conclusion}
\label{sec:discussion}

The GREEN improvements we report on three radiology VLMs, are statistically significant but moderate in absolute terms compared with what task-specific fine-tuning or RL delivers.
% : $+3.9\%$ relative on RadVLM, $+10.2\%$ on LLaVA-Rad, $+11.7\%$ on CheXOne (Table~\ref{tab:main_results})
We therefore do not present the method as a replacement for fine-tuning.
Instead, the main contribution of our approach is providing a way to introspect and open parts of the model's black box.
With per-layer SAEs and causal feature screening, as \emph{a mechanistic window} into a medical VLM's residual stream, a medical VLM can be described in terms of a few hundred sparse coordinates whose effects can be named, located in the report, and edited at inference time.
This approach avoids any weight updates and substantially reduces GPU-hour costs per model compared with post-training or RL.
The GREEN gains show that these features are not only interpretable; they are also faithful enough to steer.
\paragraph{Hallucination has type-specific structure, not a single "hallucination direction".}
The feature analysis shows that hallucination is not one global direction that can simply be turned down.
The same SAE coordinate can help one GREEN error type while hurting another: for example, \#1925 appears in MF-boost and WL-suppress, and \#30874 appears in MF-suppress and FF-boost (Table~\ref{tab:feature_pertype_summary_L16}).
At $\ell{=}16$, $80/500$ screened coordinates simultaneously satisfy $\Delta_{\mathrm{FF}}<0$ and $\Delta_{\mathrm{MF}}>0$.
This is why we keep separate suppress and boost lists for each error type.
The structure also changes across layers: $\ell{=}24$ is almost entirely FF-boost ($494/500$), while wrong-severity features move from mixed at $\ell{=}8$ to strongly WS-suppress at $\ell{=}16$ ($450$ vs.\ $14$).
This kind of per-error, per-layer organization is hidden by aggregate scores.
\paragraph{Trade-offs are localizable and transparent.}
The main trade-off is also interpretable.
Combined steering recovers $654$ missing findings but adds $1,112$ false findings (Table~\ref{tab:per_error}), which could look like a noisier model.
The SAE features show a more specific story.
One sparse, late-report feature (\#15810; $43/9,951$ studies, $87\%$ in the late half) is activated on repeated phrases, and two dense features (\#5872, \#16965) fire on generic normal-report templates.
On the other side, features \#2541 and \#4240 activate on real findings and end-of-report negation lists (Tables~\ref{tab:feature_examples_L16},~\ref{feature_profile_L16}).
The trade-off is therefore not random fabrication; it is more complete reporting with some templated repetition.
This also explains why GREEN, RadGraph, CheXbert, and RadCliQ improve together despite the FF increase: recovering a missing finding has a larger effect under GREEN's structure than adding one denominator-only error (Eq.~\ref{eq:green}).
Because the edit is controlled by $\alpha$ and by per-error-type lists, a user who needs a stricter FF limit can adjust $\alpha$ on a local validation set.
\paragraph{What transfers across models, and what does not.}
The cross-model alignment study shows a simple message.
Boost features look similar across architectures (signature cosine $0.91$-$0.95$), but suppress features are much more model-specific (as low as $0.25\,[0.12,0.40]$ for LLaVA-Rad vs.\ CheXOne; Table~\ref{tab:census}).
In other words, models share some quality-promoting directions, but each model develops its own hallucination pathways.
This matters for transfer: boost priors may transfer, while suppress features should be re-examined for each backbone.
The largest gain appears on CheXOne, even though it is already GRPO-tuned, suggesting that SAE steering and RL are complementary rather than redundant.

As an \textbf{outlook}, our mechanistic study opens concrete follow-up work that aggregate metric evaluation alone does not suggest: census-aligned steering recipes that share boost priors across models while specialising suppress weights per backbone; additional inference-time baselines (ITI~\citep{li2023inferencetime}, DoLa\citep{chuang2024dola}) to compare with other lightweight editing methods; joint language-stream and vision-stream steering \citep{pach2026sparse}; and extension beyond chest radiography.
More broadly, we hope that demonstrating \emph{interpretable} hallucination mitigation,  where every edit maps to a named SAE feature, a measurable causal effect, and a localisable position in the generated report, encourages the field to prioritize mechanistic transparency alongside advances in evaluation metrics.

\section*{Limitations}
Our study is offline, on MIMIC-CXR and IU-Xray English reports; clinical deployment requires regulatory scrutiny well beyond the scope of this paper.
GREEN and the other automatic metrics we use are proxies for radiologist judgment; the GREEN error taxonomy may also miss important clinical distinctions. 
We exclude comparison-type errors (false/missing comparison) because the subset of MIMIC-CXR that we employed does not include prior studies for comparison.
Generalization beyond chest radiography is an open question. Because of MIMIC's data usage agreement, we provide our quantitative examples using the IU-Xray dataset.
Table~\ref{tab:main_results} applies the \emph{same} SAE steering protocol to \emph{RadVLM, LLaVA-Rad, and CheXOne} (three architectures: two supervised, one GRPO-tuned), each with its own SAE and a one-shot test evaluation; further backbones (e.g., other Qwen-family models) and census-driven steering variants beyond the validated Combined recipe are left to future work.
The SAE dictionary size and sparsity introduce hyperparameters whose interactions with model scale are not fully characterized.
Finally, we intervene on the LLM's residual stream only and not on the vision encoder.
\citet{pach2026sparse} shows that vision-stream SAEs can also steer multimodal output without language-model edits.
A natural extension is to combine the two intervention sites for medical VLMs, using language-stream features to suppress hallucinated phrasing (this work) and vision-stream features to amplify or attenuate radiographic concepts at an earlier stage, and we leave that combination to future work.
\section*{Ethics Statement}
This work uses the publicly available MIMIC-CXR-JPG dataset under its established data-use agreement and the IU-Xray dataset.
The method is intended for research and is not validated for clinical decision-making.
Automated report-generation systems, steered or unsteered ones, should not replace radiologist review in clinical practice.
Any deployment would require prospective validation and regulatory approval.

We used an LLM assistant for limited rewriting and paraphrasing of parts of the paper; all experiments, analyses, and final claims in this manuscript were produced and verified by the authors.

\section*{Acknowledgments}
This work was supported by the Swiss AI Initiative through a grant from the Swiss National Supercomputing Centre (CSCS), under project ID a135 on Alps. Additional support was provided by the Swiss National Science Foundation (SNSF) under grant 10003518, as well as by RADICAL (Project-Call 2024.1, ID: 9), funded by the Digital Society Initiative Zurich (DIZH).

% Bibliography entries for the entire Anthology, followed by custom entries
%\bibliography{anthology,custom}
% Custom bibliography entries only
\bibliography{custom}

@article{bricken2023towards,
  title={Towards monosemanticity: Decomposing language models with dictionary learning},
  author={Bricken, Trenton and Templeton, Adly and Batson, Joshua and Chen, Brian and Jermyn, Adam and Conerly, Tom and Turner, Nick and Anil, Cem and Denison, Carson and Askell, Amanda and others},
  journal={Transformer Circuits Thread},
  volume={2},
  number={5},
  pages={6},
  year={2023}
}

@article{deperrois2025radvlminstructiondataset,
  author = {Deperrois, Nicolas and Matsuo, Hidetoshi and Ruiperez-Campillo, Samuel and Vandenhirtz, Moritz and Laguna, Sonia and Ryser, Alain and Fujimoto, Koji and Nishio, Mizuho and Sutter, Thomas and Vogt, Julia and Kluckert, Jonas and Frauenfelder, Thomas and Bluethgen, Christian and Nooralahzadeh, Farhad and Krauthammer, Michael},
  title = {{RadVLM Instruction Dataset}},
  journal = {{PhysioNet}},
  year = {2025},
  month = sep,
  note = {Version 1.0.0},
  doi = {10.13026/et5g-h222},
  url = {https://doi.org/10.13026/et5g-h222}
}

@inproceedings{
huben2024sparse,
title={Sparse Autoencoders Find Highly Interpretable Features in Language Models},
author={Robert Huben and Hoagy Cunningham and Logan Riggs Smith and Aidan Ewart and Lee Sharkey},
booktitle={The Twelfth International Conference on Learning Representations},
year={2024},
url={https://openreview.net/forum?id=F76bwRSLeK}
}

@inproceedings{
gao2025scaling,
title={Scaling and evaluating sparse autoencoders},
author={Leo Gao and Tom Dupre la Tour and Henk Tillman and Gabriel Goh and Rajan Troll and Alec Radford and Ilya Sutskever and Jan Leike and Jeffrey Wu},
booktitle={The Thirteenth International Conference on Learning Representations},
year={2025},
url={https://openreview.net/forum?id=tcsZt9ZNKD}
}

@article{templeton2024scaling,
       title={Scaling Monosemanticity: Extracting Interpretable Features from Claude 3 Sonnet},
       author={Templeton, Adly and Conerly, Tom and Marcus, Jonathan and Lindsey, Jack and Bricken, Trenton and Chen, Brian and Pearce, Adam and Citro, Craig and Ameisen, Emmanuel and Jones, Andy and Cunningham, Hoagy and Turner, Nicholas L and McDougall, Callum and MacDiarmid, Monte and Freeman, C. Daniel and Sumers, Theodore R. and Rees, Edward and Batson, Joshua and Jermyn, Adam and Carter, Shan and Olah, Chris and Henighan, Tom},
       year={2024},
       journal={Transformer Circuits Thread},
       url={https://transformer-circuits.pub/2024/scaling-monosemanticity/index.html}
    }

@inproceedings{
marks2025sparse,
title={Sparse Feature Circuits: Discovering and Editing Interpretable Causal Graphs in Language Models},
author={Samuel Marks and Can Rager and Eric J Michaud and Yonatan Belinkov and David Bau and Aaron Mueller},
booktitle={The Thirteenth International Conference on Learning Representations},
year={2025},
url={https://openreview.net/forum?id=I4e82CIDxv}
}

@inproceedings{
postmus2024steering,
title={Steering Large Language Models using Conceptors: Improving Addition-Based Activation Engineering},
author={Joris Postmus and Steven Abreu},
booktitle={MINT: Foundation Model Interventions},
year={2024},
url={https://openreview.net/forum?id=gyAnAq16HC}
}

@misc{zou2023transparency,
      title={Representation Engineering: A Top-Down Approach to AI Transparency}, 
      author={Andy Zou and Long Phan and Sarah Chen and James Campbell and Phillip Guo and Richard Ren and Alexander Pan and Xuwang Yin and Mantas Mazeika and Ann-Kathrin Dombrowski and Shashwat Goel and Nathaniel Li and Michael J. Byun and Zifan Wang and Alex Mallen and Steven Basart and Sanmi Koyejo and Dawn Song and Matt Fredrikson and Zico Kolter and Dan Hendrycks},
      year={2023},
      eprint={2310.01405},
      archivePrefix={arXiv},
      primaryClass={cs.CL}
}

@inproceedings{
li2023inferencetime,
title={Inference-Time Intervention: Eliciting Truthful Answers from a Language Model},
author={Kenneth Li and Oam Patel and Fernanda Vi{\'e}gas and Hanspeter Pfister and Martin Wattenberg},
booktitle={Thirty-seventh Conference on Neural Information Processing Systems},
year={2023},
url={https://openreview.net/forum?id=aLLuYpn83y}
}

@inproceedings{arad-etal-2025-saes,
    title = "{SAE}s Are Good for Steering {--} If You Select the Right Features",
    author = "Arad, Dana  and
      Mueller, Aaron  and
      Belinkov, Yonatan",
    editor = "Christodoulopoulos, Christos  and
      Chakraborty, Tanmoy  and
      Rose, Carolyn  and
      Peng, Violet",
    booktitle = "Proceedings of the 2025 Conference on Empirical Methods in Natural Language Processing",
    month = nov,
    year = "2025",
    address = "Suzhou, China",
    publisher = "Association for Computational Linguistics",
    url = "https://aclanthology.org/2025.emnlp-main.519/",
    doi = "10.18653/v1/2025.emnlp-main.519",
    pages = "10241--10259",
    ISBN = "979-8-89176-332-6",

    
}

@inproceedings{
o2024steering,
title={Steering Language Model Refusal with Sparse Autoencoders},
author={Kyle O'Brien and David Majercak and Xavier Fernandes and Richard G. Edgar and Blake Bullwinkel and Jingya Chen and Harsha Nori and Dean Carignan and Eric Horvitz and Forough Poursabzi-Sangdeh},
booktitle={ICML 2025 Workshop on Reliable and Responsible Foundation Models},
year={2025},
url={https://openreview.net/forum?id=PMK1jdGQoc}
}

@inproceedings{
pach2026sparse,
title={Sparse Autoencoders Learn Monosemantic Features in Vision-Language Models},
author={Mateusz Pach and Shyamgopal Karthik and Quentin Bouniot and Serge Belongie and Zeynep Akata},
booktitle={The Thirty-ninth Annual Conference on Neural Information Processing Systems},
year={2026},
url={https://openreview.net/forum?id=DaNnkQJSQf}
}

@inproceedings{ostmeier-etal-2024-green,
    title = "{GREEN}: Generative Radiology Report Evaluation and Error Notation",
    author = "Ostmeier, Sophie  and
      Xu, Justin  and
      Chen, Zhihong  and
      Varma, Maya  and
      Blankemeier, Louis  and
      Bluethgen, Christian  and
      Michalson, Arne Edward  and
      Moseley, Michael  and
      Langlotz, Curtis  and
      Chaudhari, Akshay S  and
      Delbrouck, Jean-Benoit",
    editor = "Al-Onaizan, Yaser  and
      Bansal, Mohit  and
      Chen, Yun-Nung",
    booktitle = "Findings of the Association for Computational Linguistics: EMNLP 2024",
    month = nov,
    year = "2024",
    address = "Miami, Florida, USA",
    publisher = "Association for Computational Linguistics",
    url = "https://aclanthology.org/2024.findings-emnlp.21/",
    doi = "10.18653/v1/2024.findings-emnlp.21",
    pages = "374--390",
}

@inproceedings{xu-etal-2025-radeval,
    title = "{R}ad{E}val: A framework for radiology text evaluation",
    author = "Xu, Justin  and
      Zhang, Xi  and
      Abderezaei, Javid  and
      Bauml, Julie  and
      Boodoo, Roger  and
      Haghighi, Fatemeh  and
      Ganjizadeh, Ali  and
      Brattain, Eric  and
      Van Veen, Dave  and
      Meng, Zaiqiao  and
      Eyre, David W  and
      Delbrouck, Jean-Benoit",
    editor = {Habernal, Ivan  and
      Schulam, Peter  and
      Tiedemann, J{\"o}rg},
    booktitle = "Proceedings of the 2025 Conference on Empirical Methods in Natural Language Processing: System Demonstrations",
    month = nov,
    year = "2025",
    address = "Suzhou, China",
    publisher = "Association for Computational Linguistics",
    url = "https://aclanthology.org/2025.emnlp-demos.40/",
    doi = "10.18653/v1/2025.emnlp-demos.40",
    pages = "546--557",
    ISBN = "979-8-89176-334-0",
}

@inproceedings{
jain2021radgraph,
title={RadGraph: Extracting Clinical Entities and Relations from Radiology Reports},
author={Saahil Jain and Ashwin Agrawal and Adriel Saporta and Steven Truong and Du Nguyen Duong and Tan Bui and Pierre Chambon and Yuhao Zhang and Matthew P. Lungren and Andrew Y. Ng and Curtis Langlotz and Pranav Rajpurkar},
booktitle={Thirty-fifth Conference on Neural Information Processing Systems Datasets and Benchmarks Track (Round 1)},
year={2021},
url={https://openreview.net/forum?id=pMWtc5NKd7V}
}

@inproceedings{smit-etal-2020-combining,
    title = "Combining Automatic Labelers and Expert Annotations for Accurate Radiology Report Labeling Using {BERT}",
    author = "Smit, Akshay  and
      Jain, Saahil  and
      Rajpurkar, Pranav  and
      Pareek, Anuj  and
      Ng, Andrew  and
      Lungren, Matthew",
    editor = "Webber, Bonnie  and
      Cohn, Trevor  and
      He, Yulan  and
      Liu, Yang",
    booktitle = "Proceedings of the 2020 Conference on Empirical Methods in Natural Language Processing (EMNLP)",
    month = nov,
    year = "2020",
    address = "Online",
    publisher = "Association for Computational Linguistics",
    url = "https://aclanthology.org/2020.emnlp-main.117/",
    doi = "10.18653/v1/2020.emnlp-main.117",
    pages = "1500--1519",
}

@article{deperrois2025radvlm,
  title={RadVLM: A Multitask Conversational Vision-Language Model for Radiology},
  author={Nicolas Deperrois and Hidetoshi Matsuo and Samuel Ruip{\'e}rez-Campillo and Moritz Vandenhirtz and Sonia Laguna and Alain Ryser and Koji Fujimoto and Mizuho Nishio and Thomas M. Sutter and Julia E. Vogt and Jonas Kluckert and Thomas Frauenfelder and Christian Bl{\"u}thgen and Farhad Nooralahzadeh and Michael Krauthammer},
  journal={ArXiv},
  year={2025},
  volume={abs/2502.03333},
  url={https://api.semanticscholar.org/CorpusID:276116530}
}

@article{PhysioNet-mimic-cxr-2.1.0,
  author = {Johnson, Alistair and Pollard, Tom and Mark, Roger and Berkowitz, Seth and Horng, Steven},
  title = {{MIMIC-CXR Database}},
  journal = {{PhysioNet}},
  year = {2024},
  month = jul,
  note = {Version 2.1.0},
  doi = {10.13026/4jqj-jw95},
  url = {https://doi.org/10.13026/4jqj-jw95}
}

@inproceedings{Zhang2020BERTScore,
title={BERTScore: Evaluating Text Generation with BERT},
author={Tianyi Zhang and Varsha Kishore and Felix Wu and Kilian Q. Weinberger and Yoav Artzi},
booktitle={International Conference on Learning Representations},
year={2020},
url={https://openreview.net/forum?id=SkeHuCVFDr}
}

@article {Yu2022.08.30.22279318,
	author = {Yu, Feiyang and Endo, Mark and Krishnan, Rayan and Pan, Ian and Tsai, Andy and Reis, Eduardo Pontes and Fonseca, Eduardo Kaiser Ururahy Nunes and Ho Lee, Henrique Min and Abad, Zahra Shakeri Hossein and Ng, Andrew Y. and Langlotz, Curtis P. and Venugopal, Vasantha Kumar and Rajpurkar, Pranav},
	title = {Evaluating Progress in Automatic Chest X-Ray Radiology Report Generation},
	elocation-id = {2022.08.30.22279318},
	year = {2022},
	doi = {10.1101/2022.08.30.22279318},
	publisher = {Cold Spring Harbor Laboratory Press},
	URL = {https://www.medrxiv.org/content/early/2022/08/31/2022.08.30.22279318},
	eprint = {https://www.medrxiv.org/content/early/2022/08/31/2022.08.30.22279318.full.pdf},
	journal = {medRxiv}
}

@misc{zhang2026reasoning,
      title={A Reasoning-Enabled Vision-Language Foundation Model for Chest X-ray Interpretation}, 
      author={Yabin Zhang and Chong Wang and Yunhe Gao and Jiaming Liu and Maya Varma and Justin Xu and Sophie Ostmeier and Jin Long and Sergios Gatidis and Seena Dehkharghani and Arne Michalson and Eun Kyoung Hong and Christian Bluethgen and Haiwei Henry Guo and Alexander Victor Ortiz and Stephan Altmayer and Sandhya Bodapati and Joseph David Janizek and Ken Chang and Jean-Benoit Delbrouck and Akshay S. Chaudhari and Curtis P. Langlotz},
      year={2026},
      eprint={2604.00493},
      archivePrefix={arXiv},
      primaryClass={cs.CV},
      url={https://arxiv.org/abs/2604.00493}, 
}

@inproceedings{
gundersen2026radvlmgrpo,
title={Rad{VLM}-{GRPO}: Enhancing Chest X-ray Report Generation and Visual Grounding via Reinforcement Learning},
author={Benjamin Gundersen and Nicolas Deperrois and Samuel Ruiperez-Campillo and Thomas M. Sutter and Julia E Vogt and Michael Moor and Farhad Nooralahzadeh and Michael Krauthammer},
booktitle={Medical Imaging with Deep Learning},
year={2026},
url={https://openreview.net/forum?id=oVQmF3ncf0}
}

@misc{zhang2026reasoningenabledvisionlanguagefoundationmodel,
      title={A Reasoning-Enabled Vision-Language Foundation Model for Chest X-ray Interpretation}, 
      author={Yabin Zhang and Chong Wang and Yunhe Gao and Jiaming Liu and Maya Varma and Justin Xu and Sophie Ostmeier and Jin Long and Sergios Gatidis and Seena Dehkharghani and Arne Michalson and Eun Kyoung Hong and Christian Bluethgen and Haiwei Henry Guo and Alexander Victor Ortiz and Stephan Altmayer and Sandhya Bodapati and Joseph David Janizek and Ken Chang and Jean-Benoit Delbrouck and Akshay S. Chaudhari and Curtis P. Langlotz},
      year={2026},
      eprint={2604.00493},
      archivePrefix={arXiv},
      primaryClass={cs.CV},
      url={https://arxiv.org/abs/2604.00493}, 
}

@Article{ZambranoChaves2025,
author={Zambrano Chaves, Juan Manuel and Huang, Shih-Cheng and Xu, Yanbo and Xu, Hanwen and Usuyama, Naoto and Zhang, Sheng and Wang, Fei and Xie, Yujia and Khademi, Mahmoud and Yang, Ziyi and Awadalla, Hany and Gong, Julia and Hu, Houdong and Yang, Jianwei and Li, Chunyuan and Gao, Jianfeng and Gu, Yu and Wong, Cliff and Wei, Mu and Naumann, Tristan and Chen, Muhao and Lungren, Matthew P. and Chaudhari, Akshay and Yeung-Levy, Serena and Langlotz, Curtis P. and Wang, Sheng and Poon, Hoifung},
title={A clinically accessible small multimodal radiology model and evaluation metric for chest X-ray findings},
journal={Nature Communications},
year={2025},
month={Apr},
day={01},
volume={16},
number={1},
pages={3108},
issn={2041-1723},
doi={10.1038/s41467-025-58344-x},
url={https://doi.org/10.1038/s41467-025-58344-x}
}

@inproceedings{tanida2023interactive,
    title={Interactive and Explainable Region-guided Radiology Report Generation},
    author={Tanida, Tim and Müller, Philip and Kaissis, Georgios and Rueckert, Daniel},
    booktitle={CVPR},
    year={2023}
}

@inproceedings{leng2024mitigating,
  title={Mitigating object hallucinations in large vision-language models through visual contrastive decoding},
  author={Leng, Sicong and Zhang, Hang and Chen, Guanzheng and Li, Xin and Lu, Shijian and Miao, Chunyan and Bing, Lidong},
  booktitle={Proceedings of the IEEE/CVF Conference on Computer Vision and Pattern Recognition},
  pages={13872--13882},
  year={2024}
}

@inproceedings{huang2024opera,
  title={Opera: Alleviating hallucination in multi-modal large language models via over-trust penalty and retrospection-allocation},
  author={Huang, Qidong and Dong, Xiaoyi and Zhang, Pan and Wang, Bin and He, Conghui and Wang, Jiaqi and Lin, Dahua and Zhang, Weiming and Yu, Nenghai},
  booktitle={Proceedings of the IEEE/CVF Conference on Computer Vision and Pattern Recognition},
  pages={13418--13427},
  year={2024}
}

@inproceedings{chuang2024dola,
  title={Dola: Decoding by contrasting layers improves factuality in large language models},
  author={Chuang, Yung-Sung and Xie, Yujia and Luo, Hongyin and Kim, Yoon and Glass, James R and He, Pengcheng},
  booktitle={International Conference on Learning Representations},
  volume={2024},
  pages={54158--54183},
  year={2024}
}

@inproceedings{hein-etal-2025-chexalign,
    title = "{C}he{X}align: Preference fine-tuning in chest {X}-ray interpretation models without human feedback",
    author = "Hein, Dennis  and
      Chen, Zhihong  and
      Ostmeier, Sophie  and
      Xu, Justin  and
      Varma, Maya  and
      Reis, Eduardo Pontes  and
      Michalson, Arne Edward  and
      Bluethgen, Christian  and
      Shin, Hyun Joo  and
      Langlotz, Curtis  and
      Chaudhari, Akshay S",
    editor = "Che, Wanxiang  and
      Nabende, Joyce  and
      Shutova, Ekaterina  and
      Pilehvar, Mohammad Taher",
    booktitle = "Proceedings of the 63rd Annual Meeting of the Association for Computational Linguistics (Volume 1: Long Papers)",
    month = jul,
    year = "2025",
    address = "Vienna, Austria",
    publisher = "Association for Computational Linguistics",
    url = "https://aclanthology.org/2025.acl-long.1342/",
    doi = "10.18653/v1/2025.acl-long.1342",
    pages = "27679--27702",
    ISBN = "979-8-89176-251-0"
}

@article{calamida2023radiology,
  title={Radiology-aware model-based evaluation metric for report generation},
  author={Calamida, Amos and Nooralahzadeh, Farhad and Rohanian, Morteza and Fujimoto, Koji and Nishio, Mizuho and Krauthammer, Michael},
  journal={arXiv preprint arXiv:2311.16764},
  year={2023}
}

@techreport{Bannur2024-ek,
author = {Bannur, Shruthi and Bouzid, Kenza and Coelho de Castro, Daniel and Schwaighofer, Anton and Bond-Taylor, Sam and Ilse, Maximilian and Pérez-García, Fernando and Salvatelli, Valentina and Sharma, Harshita and Meissen, Felix and Ranjit, Mercy and Srivastav, Shaury and Gong, Julia and Falck, Fabian and Oktay, Ozan and Thieme, Anja and Lungren, Matthew P and Wetscherek, Maria Teodora and Alvarez-Valle, Javier and Hyland, Stephanie},
title = {MAIRA-2: Grounded Radiology Report Generation},
institution = {Microsoft},
year = {2024},
month = {June},
url = {https://www.microsoft.com/en-us/research/publication/maira-2-grounded-radiology-report-generation/},
number = {MSR-TR-2024-18},
}

@article{demner2016preparing,
  title={Preparing a collection of radiology examinations for distribution and retrieval},
  author={Demner-Fushman, Dina and others},
  journal={Journal of the American Medical Informatics Association},
  volume={23},
  number={2},
  pages={304--310},
  year={2016},
  publisher={Oxford University Press}
}

\newpage
\section*{Appendix}
\newcommand{\AppendixQualCXR}[1]{%
  \begingroup
  \setlength{\fboxsep}{3pt}%
  \setlength{\fboxrule}{0.6pt}%
  \IfFileExists{figures/appendix_qual/#1.jpg}{%
    \includegraphics[width=\linewidth,height=0.22\textheight,keepaspectratio]{figures/appendix_qual/#1.jpg}%
  }{%
    \IfFileExists{figures/appendix_qual/#1.png}{%
      \includegraphics[width=\linewidth,height=0.22\textheight,keepaspectratio]{figures/appendix_qual/#1.png}%
    }{%
      \fcolorbox{black!25}{black!4}{%
        \parbox[c][2.4cm][c]{0.98\linewidth}{%
          \centering\footnotesize\sffamily
          \textbf{CXR.}\;
          Add \texttt{appendix\_qual/#1.jpg} or \texttt{.png}.%
        }%
      }%
    }%
  }%
  \endgroup
}

% Appendix qualitative table: vertical case label; GREEN-style error tags on generated text.
\newlength{\AppendixQualRowHeight}
\setlength{\AppendixQualRowHeight}{0.19\textheight}
% First column: plain label, rotated 90° (read upward).
\newcommand{\AppendixQualCaseVert}[1]{%
  \parbox[c][\AppendixQualRowHeight]{0.03\textwidth}{\centering\vfill\rotatebox{90}{\small\sffamily\bfseries #1}\vfill}}
% Small tags after phrases (GREEN taxonomy abbrev.; see table footnote in caption).
\newcommand{\ApqGerr}[1]{\,\raisebox{0.15ex}{\scriptsize\sffamily\textcolor{red!72!black}{\textbf{[#1]}}}}
\newcommand{\ApqGwarn}[1]{\,\raisebox{0.15ex}{\scriptsize\sffamily\textcolor{orange!70!black}{\textbf{[#1]}}}}
\newcommand{\ApqGok}[1]{\,\raisebox{0.15ex}{\scriptsize\sffamily\textcolor{teal!45!black}{\textbf{[#1]}}}}
\newcommand{\ApqGinsig}[1]{\,\raisebox{0.15ex}{\scriptsize\sffamily\textcolor{gray!55!black}{\textbf{[#1]}}}}

\appendix

\section{Layer selection}
\label{sec:appendix_layer_selection}

We train SAEs at ten evenly-spaced depths $\{0,4,8,12,16,20,24,28,32,35\}$ of RadVLM's 36-layer Qwen3-VL backbone but hook only a subset $\mathcal{L}{=}\{8,16,20,24\}$ for steering.
Table~\ref{tab:layer_quality} reports SAE reconstruction quality at each depth, revealing three regimes that motivate this choice.

\paragraph{Early layers ($\ell \le 4$): poor decomposition.}
The SAE struggles to reconstruct these activations faithfully.
Early transformer layers mainly encode positional and syntactic information rather than clinical content~\citep{templeton2024scaling}, so there is little to steer.

\paragraph{Late layers ($\ell \ge 28$) : norms too large.}
Reconstruction quality is fine, but activation norms explode ($244$--$1{,}412$ vs.\ $31$--$128$ in the steered range).
Our steering perturbation $\delta$ is a fixed-size nudge; when the residual stream is $10$--$45\times$ larger, that nudge becomes negligible unless $\alpha$ is cranked up to values that break generation.
These layers are also close to the output and mainly encode next-token predictions, not factual content~\citep{gao2025scaling,bricken2023towards}.

\paragraph{Middle-to-late layers ($8 \le \ell \le 24$) : the sweet spot.}
Cosine similarity is high ($>0.97$) and norms stay moderate, so the SAE can both decompose and meaningfully edit the hidden states.
We pick $\{8,16,20,24\}$ as roughly quarter-spaced depths through this range, consistent with where prior work finds semantic and factual features in LLMs~\citep{templeton2024scaling,marks2025sparse}.
Layer $12$ has comparable quality and could be added; we leave it out to keep the causal screening budget tractable (${\sim}5$ extra GPU-h per layer).
We observe that a multi-layer ablation in \S\ref{grid_search} confirms that steering all four layers outperforms any single-layer or reduced subset.

\begin{table}[t]
  \centering
  \small
  \begin{tabular}{@{}rccr@{}}
    \toprule
    \textbf{Layer} & \textbf{Cos.\ sim.} $\uparrow$ & \textbf{Dead \%} $\downarrow$ & \textbf{$\|\mathbf{h}\|$} \\
    \midrule
    0  & 0.993 & 63.9 & 8 \\
    4  & 0.992 & 78.0 & 22 \\
    \midrule
    \textbf{8}  & 0.985 & 77.3 & 31 \\
    12 & 0.981 & 81.7 & 38 \\
    \textbf{16} & 0.982 & 81.9 & 49 \\
    \textbf{20} & 0.974 & 82.6 & 64 \\
    \textbf{24} & 0.969 & 84.6 & 128 \\
    \midrule
    28 & 0.970 & 83.1 & 244 \\
    32 & 0.980 & 82.5 & 510 \\
    35 & 0.994 & 82.4 & 1412 \\
    \bottomrule
  \end{tabular}
  \caption{\textbf{SAE reconstruction quality across all ten trained layers} (RadVLM, $D{=}32{,}768$, $k{=}64$).
  \emph{Cos.\ sim.}: mean cosine similarity between input and reconstruction ($>0.97$ for all steered layers).
  \emph{Dead \%}: fraction of dictionary features never activated.
  $\|\mathbf{h}\|$: mean activation norm.
  \textbf{Bold} layer numbers mark the steered set $\mathcal{L}{=}\{8,16,20,24\}$.
  Horizontal rules separate three regimes:
  early layers ($0$--$4$) with poor feature decomposition;
  middle-to-late layers ($8$--$24$) with high reconstruction quality and moderate norms;
  late layers ($28$--$35$) where norms grow $8$--$45\times$, making steering perturbations relatively small.}
  \label{tab:layer_quality}
\end{table}

\section{Compute}
Table~\ref{tab:efficiency} compares the compute cost of SAE steering against GRPO fine-tuning.
The full SAE pipeline, activation collection through causal feature identification, takes roughly $27$ GPU-hours on NVIDIA GH200 for a single model (RadVLM); SAE training itself is $<$\,1 GPU-hour.
GRPO fine-tuning takes multiple days of RL training per model.
At inference time, SAE steering is essentially free: one SAE encode/decode per layer per token, on tensors already in GPU memory ($<$5\% wall-clock overhead).
\begin{table}[t]
  \centering
  \footnotesize
  \setlength{\tabcolsep}{4pt}
  \resizebox{\columnwidth}{!}{%
  \begin{tabular}{@{}lcc@{}}
    \toprule
    \textbf{Component} & \textbf{SAE steering} & \textbf{GRPO} \\
    \midrule
    Requires training?             & No (frozen weights)  & Yes \\
    Activation collection          & ${\sim}$6 GPU-h      & --- \\
    SAE training (10 layers)       & $<$1 GPU-h           & --- \\
    Causal feature identification  & ${\sim}$20 GPU-h     & --- \\
    RL training                    & ---                  & ${\sim} 350$ GPU-h \\
    \midrule
    \textbf{Total pipeline}        & \textbf{${\sim}$27 GPU-h} & \textbf{${\sim}350$ GPU-h} \\
    Inference overhead             & $<$5\%               & 0\% \\
    Per-model retraining needed?   & Yes (SAEs)           & Yes (full RL) \\
    Interpretable features?        & \ding{51}            & \ding{55} \\
    Per-error-type control?        & \ding{51}            & \ding{55} \\
    \bottomrule
  \end{tabular}%
  }
  \caption{Compute and capability comparison between SAE steering and GRPO fine-tuning.
  GPU-hours for SAE work are on NVIDIA GH200 (96GB).
  The bulk of SAE steering's cost is the causal feature-identification stage (single-feature forward screens, ${\sim}$27 GPU-h total across all four steered layers, sharded over 4 GPUs).
  The GRPO column is the RL training cost reported by RadVLM-GRPO \citep{gundersen2026radvlmgrpo}}.
  \label{tab:efficiency}
\end{table}

\section{Extended error-redistribution analysis}
\label{sec:appendix_error_redistribution}

This section provides the extended discussion of the error-redistribution trade-off summarised in \S\ref{Per-error-type}.

\paragraph{GREEN's structural asymmetry.}
Let $M$ be the number of matched findings in a report and $E_i$ the count of significant errors of type $i\in\{$(a) FF, (b) MF, (c) WL, (d) WS, (e) false comparison, (f) missing comparison$\}$.  Then
\begin{equation}
\label{eq:green}
  \mathrm{GREEN} \;=\; \frac{M}{M + \sum_{i=\mathrm{(a)}}^{\mathrm{(f)}} E_i} \;\in\; [0,1],
\end{equation}
defined as $0$ when $M{=}0$.
A single MF$\to$match conversion therefore moves GREEN by roughly twice as much as one denominator-only addition (adding $1$ to numerator \emph{and} subtracting $1$ from denominator), so the score can rise even when the total error count grows.
In our case $654$ MF errors flip to matches while the additional $+1{,}352$ FF/WL/WS errors land on the denominator-only side, and the ratio still increases.
RadGraph entity-relation F1 (recall-weighted) and RadCliQ (calibrated to radiologist preference, smaller penalty on added than on missed content) carry the same asymmetry, which is why all three metrics move together at this operating point.

\paragraph{Why the total error rise is not a uniform quality loss.}
Two facts are easy to overlook.
\textbf{First}, mean GREEN improves, and RadGraph F1, CheXbert-14, BERTScore, and RadCliQ all move in the same direction (Table~\ref{tab:main_results}).
A model trading missing findings for unrelated false findings would not improve RadGraph or CheXbert similarity.
\textbf{Second}, GREEN's six categories are clinically asymmetric: a missed \emph{pneumothorax} is more harmful than a redundantly reported "normal-variant cardiac silhouette", and per-report mean GREEN, RadGraph, and CheXbert weight recall of reference findings more than added content.

\paragraph{What do the boost features actually activate on?}
We re-encode the RadVLM per-token validation activation pool (10,000 decodes, 9,951 distinct study\_ids) through the SAE at layer~16 and aggregate token-level pre-activations.
The FF-boost set separates into two categories:
(i)~a sparse feature associated with late-stage \emph{repetition loops} (\#15810; 43/9,951 studies, 87\% in the latter half, top activations on direct phrase repetition); and
(ii)~dense template features (\#5872, \#16965) that activate across nearly all studies and reflect \emph{generic report-normal} structures.
In contrast, the MF/WL suppress examples \#2541 and \#4240 peak on paired-finding wording and end-of-report negation chains (Table~\ref{tab:feature_examples_L16}); ablating \#2541 lowers MF ($\Delta_{\mathrm{MF}}{=}{-}0.09$) while raising FF ($\Delta_{\mathrm{FF}}{=}{+}0.14$), and ablating \#4240 lowers WL ($\Delta_{\mathrm{WL}}{=}{-}0.09$).
Together, these observations suggest that the $+1{,}112$ FF increase is better interpreted as a consequence of more complete late-report generation, rather than evidence of safety-critical fabrication.
\emph{To our knowledge, this is the first quantitative localisation of SAE features in a medical VLM.}

\paragraph{Layer-wise heterogeneity details.}
Among the screened features at $\ell{=}24$, almost every coordinate is FF-boost signed ($494/500$ with $\Delta_{\mathrm{FF}}>0$), whereas wrong-severity at $\ell{=}8$ still mixes WS-suppress and WS-boost coordinates ($234$ vs.\ $184$), but at $\ell{=}16$ this distribution becomes highly skewed toward suppression ($450$ vs.\ $14$).
Our ablation results further support the multi-layer design: removing any individual layer from $\mathcal{L}{=}\{8,16,20,24\}$ leads to a measurable degradation rather than being compensated by the remaining layers.

\section{Interpretation: Features, behaviour, and cases}
\label{sec:appendix_pertype}

\begin{figure}[t]

    \centering
    \includegraphics[width=1\linewidth]{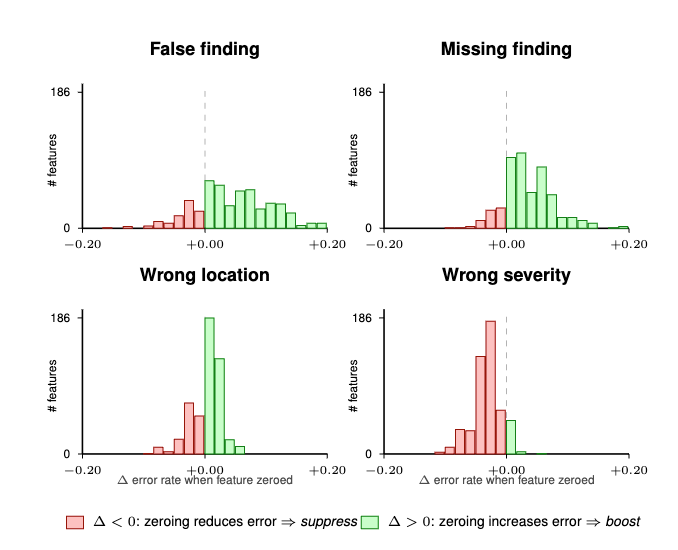}
    \caption{
\textbf{Distribution of per-error-type causal effects at layer~16 (RadVLM, 500 prefiltered features).} Each panel shows the histogram of $\Delta_{\text{type}}{=}\text{error}_{\text{ablated}}-\text{error}_{\text{baseline}}$ for one GREEN error type. Left tail (red): suppress candidates (ablation reduces this error). Right tail (green): boost candidates. The asymmetric tails motivate the per-error-type lists used in the combined intervention: features with strong $\Delta_{\text{FF}}{<}0$ are not the same set as features with strong $\Delta_{\text{MF}}{>}0$.
}
\label{fig:feature_pertype_hist_L16}
\vspace{-0.75em}
   
\end{figure}

\begin{figure}[h]
    \centering
    \includegraphics[width=1\linewidth]{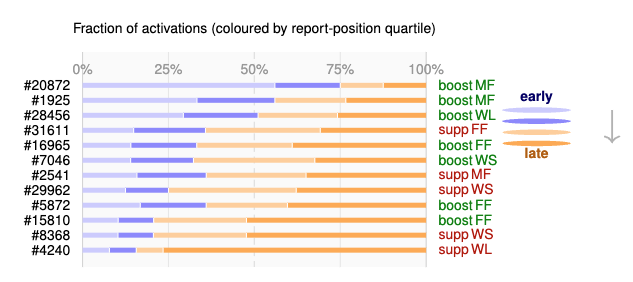}
\caption{Where each reported layer-16 feature fires in the generated report ($10{,}000$ validation decodes). Each stacked bar shows what fraction of a feature's total activations fall in each position quartile: early \textcolor{blue!5!black}{[0,\,.25)}, early-mid \textcolor{blue!45!black}{[.25,\,.50)}, late-mid \textcolor{orange!60!black}{[.50,\,.75)}, late \textcolor{orange!80!black}{[.75,\,1)}. Boost features (e.g.\ \#15810) concentrate in the late quartiles; suppress features spread more evenly. Sorted by median position (bottom = latest).}
\label{fig:feature_profile_hist_L16}
\end{figure}

Here, we walk through what the screened SAE features do in practice, in four progressively high-level subsections. We first list the strongest causal features per steered layer and the text contexts they activate on (Tables~\ref{tab:feature_pertype_top_L8}-\ref{tab:feature_pertype_top_L24} and Table~\ref{tab:feature_examples_L16}); we then quantify \emph{where} in the report each reported features activates (Table~\ref{feature_profile_L16}, Figure~\ref{fig:feature_profile_hist_L16}), we visualize the full distribution of per-error-type causal effects $\Delta_j^{(j)}$ at the canocnical layer $\ell=16$ in Figure~\ref{fig:feature_pertype_hist_L16}. A short synthesis paragraph at the end relates these observations to the cross-model boost-suppress finding in \S~\ref{sec:res_census}.

% --- layer 8 ---
\begin{table}[t]
\centering
\scriptsize
\setlength{\tabcolsep}{1.6pt}
\renewcommand{\arraystretch}{0.75}
\resizebox{\columnwidth}{!}{%
\begin{tabular}{@{}llrrrrr@{}}
\toprule
Error Type & Role & Feat. & $\Delta_{\text{FF}}$ & $\Delta_{\text{MF}}$ &
$\Delta_{\text{WL}}$ & $\Delta_{\text{WS}}$ \\
\midrule
False finding & supp. & 6596 & $-0.310$ & $+0.040$ & $-0.040$ & $-0.010$  \\
 &  & 24597 & $-0.260$ & $-0.040$ & $+0.010$ & $+0.030$  \\
 &  & 94 & $-0.240$ & $-0.020$ & $+0.040$ & $+0.000$  \\
 & boost & 18416 & $+0.200$ & $+0.060$ & $-0.040$ & $-0.020$  \\
 &  & 12906 & $+0.190$ & $+0.060$ & $-0.030$ & $-0.020$ \\
 &  & 11500 & $+0.180$ & $+0.040$ & $-0.030$ & $-0.010$  \\
\addlinespace[2pt]
Missing finding & supp. & 3983 & $-0.180$ & $-0.130$ & $+0.020$ & $+0.040$  \\
 &  & 23671 & $-0.120$ & $-0.120$ & $+0.010$ & $+0.030$  \\
 &  & 18669 & $-0.140$ & $-0.110$ & $+0.010$ & $+0.020$  \\
 & boost & 23148 & $+0.090$ & $+0.370$ & $-0.100$ & $-0.100$  \\
 &  & 1542 & $+0.150$ & $+0.250$ & $-0.100$ & $-0.070$  \\
 &  & 26996 & $-0.110$ & $+0.240$ & $-0.020$ & $+0.020$  \\
\addlinespace[2pt]
Wrong location & supp. & 13589 & $+0.070$ & $+0.200$ & $-0.130$ & $-0.040$ \\
 &  & 1542 & $+0.150$ & $+0.250$ & $-0.100$ & $-0.070$ \\
 &  & 6658 & $+0.000$ & $+0.150$ & $-0.100$ & $-0.030$  \\
 & boost & 17609 & $-0.100$ & $-0.010$ & $+0.070$ & $+0.040$  \\
 &  & 19414 & $-0.070$ & $-0.020$ & $+0.060$ & $-0.060$  \\
 &  & 19133 & $-0.050$ & $-0.020$ & $+0.050$ & $+0.000$ \\
\addlinespace[2pt]
Wrong severity & supp. & 23148 & $+0.090$ & $+0.370$ & $-0.100$ & $-0.100$\\
 &  & 21739 & $+0.010$ & $+0.100$ & $-0.060$ & $-0.080$  \\
 &  & 31838 & $+0.050$ & $-0.030$ & $-0.010$ & $-0.080$  \\
 & boost & 7210 & $-0.110$ & $-0.020$ & $+0.020$ & $+0.070$  \\
 &  & 26264 & $+0.020$ & $+0.030$ & $+0.000$ & $+0.060$ \\
 &  & 6916 & $-0.080$ & $-0.090$ & $+0.050$ & $+0.060$ \\
\bottomrule
\end{tabular}%
}
\caption{\textbf{Top causal features at layer~8.} Top-3 suppress and boost features for each GREEN error type, with deltas on FF/MF/WL/WS.}
\label{tab:feature_pertype_top_L8}
\end{table}

% --- layer 16 ---
\begin{table}[t]
\centering
\scriptsize
\setlength{\tabcolsep}{1.6pt}
\renewcommand{\arraystretch}{0.75}
\resizebox{\columnwidth}{!}{%
\begin{tabular}{@{}llrrrrrr@{}}
\toprule
Error Type & Role & Feat. & $\Delta_{\text{FF}}$ & $\Delta_{\text{MF}}$ &
$\Delta_{\text{WL}}$ & $\Delta_{\text{WS}}$  \\
\midrule
False finding & supp. & 31611 & $-0.160$ & $+0.040$ & $-0.030$ & $-0.040$ \\
 &  & 23239 & $-0.130$ & $-0.030$ & $-0.030$ & $-0.030$ \\
 &  & 8320 & $-0.130$ & $+0.100$ & $-0.070$ & $-0.050$  \\
 & boost & 15810 & $+0.300$ & $+0.080$ & $+0.000$ & $-0.020$  \\
 &  & 5872 & $+0.280$ & $+0.080$ & $+0.000$ & $-0.100$  \\
 &  & 16965 & $+0.270$ & $+0.060$ & $-0.060$ & $-0.060$  \\
\addlinespace[2pt]
Missing finding & supp. & 2541 & $+0.140$ & $-0.090$ & $-0.020$ & $-0.040$  \\
 &  & 25211 & $+0.020$ & $-0.070$ & $-0.010$ & $+0.020$ \\
 &  & 8457 & $+0.130$ & $-0.060$ & $-0.020$ & $+0.000$  \\
 & boost & 1925 & $+0.150$ & $+0.370$ & $-0.080$ & $+0.000$  \\
 &  & 20872 & $+0.110$ & $+0.230$ & $-0.020$ & $-0.080$  \\
 &  & 9114 & $+0.050$ & $+0.190$ & $-0.020$ & $-0.040$ \\
\addlinespace[2pt]
Wrong location & supp. & 4240 & $+0.200$ & $+0.060$ & $-0.090$ & $-0.030$  \\
 &  & 22773 & $+0.010$ & $+0.140$ & $-0.080$ & $-0.070$  \\
 &  & 1925 & $+0.150$ & $+0.370$ & $-0.080$ & $+0.000$  \\
 & boost & 28456 & $-0.030$ & $+0.040$ & $+0.060$ & $-0.020$  \\
 &  & 16975 & $+0.070$ & $+0.000$ & $+0.060$ & $-0.030$  \\
 &  & 11553 & $+0.080$ & $-0.020$ & $+0.050$ & $-0.010$  \\
\addlinespace[2pt]
Wrong severity & supp. & 29962 & $+0.180$ & $+0.080$ & $+0.020$ & $-0.110$ \\
 &  & 13311 & $+0.160$ & $+0.060$ & $-0.010$ & $-0.110$  \\
 &  & 8368 & $+0.220$ & $+0.170$ & $-0.030$ & $-0.100$  \\
 & boost & 7046 & $+0.040$ & $+0.010$ & $-0.040$ & $+0.050$  \\
 &  & 24384 & $+0.060$ & $+0.090$ & $+0.020$ & $+0.030$  \\
 &  & 31685 & $+0.110$ & $+0.040$ & $+0.010$ & $+0.020$ \\
\bottomrule
\end{tabular}%
}
\caption{\textbf{Top causal features at layer~16.} Top-3 suppress and boost features for each GREEN error type, with deltas on FF/MF/WL/WS.}
\label{tab:feature_pertype_top_L16}
\end{table}

% --- layer 20 ---
\begin{table}[t]
\centering
\scriptsize
\setlength{\tabcolsep}{1.6pt}
\renewcommand{\arraystretch}{0.75}
\resizebox{\columnwidth}{!}{%
\begin{tabular}{@{}llrrrrr@{}}
\toprule
Error type & Role & Feat. \# & $\Delta_{\text{FF}}$ & $\Delta_{\text{MF}}$ & $\Delta_{\text{WL}}$ & $\Delta_{\text{WS}}$  \\
\midrule
False finding & supp. & 20624 & $-0.220$ & $-0.020$ & $+0.020$ & $+0.010$   \\
 &  & 21539 & $-0.210$ & $-0.090$ & $+0.000$ & $+0.030$   \\
 &  & 22189 & $-0.210$ & $-0.020$ & $+0.010$ & $+0.000$   \\
 & boost & 30874 & $+0.330$ & $-0.140$ & $+0.030$ & $+0.090$  \\
 &  & 14416 & $+0.270$ & $+0.060$ & $-0.020$ & $+0.080$ \\
 &  & 17767 & $+0.260$ & $+0.030$ & $-0.010$ & $-0.010$  \\
\addlinespace[2pt]
Missing finding & supp. & 30623 & $+0.000$ & $-0.150$ & $+0.010$ & $+0.050$  \\
 &  & 30874 & $+0.330$ & $-0.140$ & $+0.030$ & $+0.090$  \\
 &  & 16748 & $-0.100$ & $-0.140$ & $+0.000$ & $+0.040$  \\
 & boost & 24418 & $+0.050$ & $+0.220$ & $-0.050$ & $-0.050$  \\
 &  & 16881 & $+0.090$ & $+0.170$ & $-0.010$ & $-0.030$ \\
 &  & 5385 & $+0.100$ & $+0.150$ & $-0.030$ & $-0.020$  \\
\addlinespace[2pt]
Wrong location & supp. & 9659 & $-0.090$ & $+0.030$ & $-0.110$ & $-0.010$  \\
 &  & 15819 & $-0.070$ & $+0.080$ & $-0.080$ & $-0.010$  \\
 &  & 2651 & $+0.010$ & $-0.050$ & $-0.080$ & $+0.020$  \\
 & boost & 5091 & $-0.110$ & $-0.130$ & $+0.050$ & $+0.020$  \\
 &  & 26348 & $-0.140$ & $-0.110$ & $+0.050$ & $+0.030$ \\
 &  & 29751 & $-0.060$ & $-0.090$ & $+0.050$ & $+0.030$  \\
\addlinespace[2pt]
Wrong severity & supp. & 5571 & $-0.180$ & $+0.050$ & $-0.010$ & $-0.060$ \\
 &  & 24418 & $+0.050$ & $+0.220$ & $-0.050$ & $-0.050$ \\
 &  & 22955 & $-0.040$ & $+0.070$ & $+0.000$ & $-0.040$  \\
 & boost & 30874 & $+0.330$ & $-0.140$ & $+0.030$ & $+0.090$  \\
 &  & 14416 & $+0.270$ & $+0.060$ & $-0.020$ & $+0.080$   \\
 &  & 3685 & $-0.100$ & $-0.060$ & $+0.020$ & $+0.080$  \\
\bottomrule
\end{tabular}%
}
\caption{\textbf{Top causal features at layer~20.} Top-3 suppress and boost features for each GREEN error type, with deltas on FF/MF/WL/WS.}
\label{tab:feature_pertype_top_L20}
\end{table}

% --- layer 24 ---
\begin{table}[t]
\centering
\scriptsize
\setlength{\tabcolsep}{1.6pt}
\renewcommand{\arraystretch}{0.75}
\resizebox{\columnwidth}{!}{%
\begin{tabular}{@{}llrrrrr@{}}
\toprule
Error type & Role & Feat. \# & $\Delta_{\text{FF}}$ & $\Delta_{\text{MF}}$ & $\Delta_{\text{WL}}$ & $\Delta_{\text{WS}}$  \\
\midrule
False finding & supp. & 14217 & $-0.020$ & $+0.110$ & $-0.020$ & $-0.010$  \\
 &  & 1685 & $-0.020$ & $+0.000$ & $-0.010$ & $-0.030$  \\
 &  & 10884 & $-0.020$ & $+0.000$ & $-0.010$ & $-0.040$ \\
 & boost & 20420 & $+0.560$ & $+0.010$ & $-0.070$ & $+0.090$  \\
 &  & 3026 & $+0.360$ & $+0.040$ & $-0.030$ & $-0.010$  \\
 &  & 15397 & $+0.310$ & $-0.030$ & $-0.050$ & $+0.010$  \\
\addlinespace[2pt]
Missing finding & supp. & 1972 & $+0.110$ & $-0.140$ & $-0.040$ & $+0.010$  \\
 &  & 24480 & $+0.200$ & $-0.080$ & $-0.030$ & $-0.050$  \\
 &  & 20968 & $+0.060$ & $-0.070$ & $-0.010$ & $+0.030$  \\
 & boost & 15720 & $+0.080$ & $+0.270$ & $-0.020$ & $-0.090$  \\
 &  & 16413 & $+0.200$ & $+0.150$ & $-0.040$ & $-0.050$  \\
 &  & 15291 & $+0.020$ & $+0.130$ & $-0.010$ & $+0.030$  \\
\addlinespace[2pt]
Wrong location & supp. & 28349 & $+0.130$ & $+0.080$ & $-0.110$ & $-0.010$ \\
 &  & 32155 & $+0.180$ & $+0.120$ & $-0.100$ & $-0.060$  \\
 &  & 25435 & $+0.210$ & $+0.050$ & $-0.100$ & $-0.110$  \\
 & boost & 22492 & $+0.190$ & $+0.010$ & $+0.020$ & $+0.000$  \\
 &  & 11120 & $+0.120$ & $+0.010$ & $+0.020$ & $+0.010$ \\
 &  & 25793 & $+0.160$ & $-0.020$ & $+0.020$ & $+0.010$  \\
\addlinespace[2pt]
Wrong severity & supp. & 25435 & $+0.210$ & $+0.050$ & $-0.100$ & $-0.110$ \\
 &  & 15720 & $+0.080$ & $+0.270$ & $-0.020$ & $-0.090$ \\
 &  & 12903 & $+0.180$ & $+0.090$ & $-0.030$ & $-0.080$  \\
 & boost & 20420 & $+0.560$ & $+0.010$ & $-0.070$ & $+0.090$  \\
 &  & 10254 & $+0.050$ & $+0.010$ & $-0.030$ & $+0.040$  \\
 &  & 16654 & $+0.140$ & $-0.010$ & $-0.020$ & $+0.040$ \\
\bottomrule
\end{tabular}%
}
\caption{\textbf{Top causal features at layer~24.} Top-3 suppress and boost features for each GREEN error type, with deltas on FF/MF/WL/WS.}
\label{tab:feature_pertype_top_L24}
\end{table}

\paragraph{Top causal features per steered layer.} 
Tables~\ref{tab:feature_pertype_top_L8}-\ref{tab:feature_pertype_top_L24} list, for each steered layer and each GREEN error type, top-3 suppress (most negative per-type delta) and top-3 boost (most positive per-type delta) features among the 500 prefiltered candidates from the causal selection pipeline. The same SAE feature can appear in multiple of these eight (error-type $\times$ role) sub-blocks at the same time. For example, feature \#1925 at layer~16 sits in the MF-\emph{boost} block ($\Delta_{\mathrm{MF}}=+0.37$) and in the WL-\emph{suppress} block ($\Delta_{\mathrm{WL}}=-0.08$): the same coordinate plays opposite causal roles for different clinical mistakes.

Figure~\ref{fig:feature_pertype_hist_L16} visualises the full distribution of per-type deltas at $\ell{=}16$.
The figure has \emph{four panels}, one per GREEN error type (FF / MF / WL / WS); each panel is a histogram of one number per feature.

For every feature $j$ in the $500$-feature pool at $\ell{=}16$ we measured its causal effect $\Delta_t^{(j)}$ on a \emph{single} error type $t$, the average change in that error's count per report when $j$ is zeroed on the $N{=}100$ screening slice (\S\ref{method:features}).
That is one number per feature per error type, so the panel for type $t$ contains exactly $500$ values of $\Delta_t^{(j)}$.

The \emph{x-axis} is that value $\Delta_t^{(j)}$ (mean errors per report).
The \emph{y-axis} is the number of features whose $\Delta$ falls in that x-bin as a standard histogram count, not a clinical metric.
The interpretation of the bars with different colors is as follows:
\begin{itemize}[leftmargin=*,nosep]
\setlength{\itemsep}{1pt}
\item \textbf{red bars (left of $0$):} features for which zeroing \emph{reduces} that error type , the feature was \emph{promoting} the error, so it is a \emph{suppress} candidate for that type.
\item \textbf{green bars (right of $0$):} features for which zeroing \emph{increases} that error type, the feature was \emph{preventing} the error, so it is a \emph{boost} candidate for that type.
\item \textbf{bars at $0$:} features whose ablation barely moves this error type (most of the pool, by design, the prefilter only guarantees activation-difference relevance, not per-type causal sign).
\end{itemize}

The reason that we show in the four panels is that 
the same feature shows up as a \emph{different} number in each panel, because $\Delta_t^{(j)}$ depends on the error type $t$.
A given feature can sit in the left tail of one panel and in the right tail of another at the same time, e.g., \ feature \#1925 contributes to the MF-boost (right) tail and the WL-suppress (left) tail simultaneously.
That per-feature asymmetry is exactly what forces separate suppress and boost lists \emph{per error type} in the combined intervention (Eq.~\ref{eq:edit}).

\paragraph{Features are identified by causal effect, not by meaning.}

We do \emph{not} hand-label individual SAE features with clinical concepts.
A feature's identity throughout this paper is its causal effect on each error type ($\Delta_t^{(j)}$, \S\ref{method:features}), not a name or a lexical definition.
Table~\ref{tab:feature_examples_L16} is a complementary, \emph{observational} diagnostic: for each reported feature, it shows the unsteered text that activates that feature most strongly.

We generated $10,000$ validation reports with the unsteered RadVLM (\emph{no SAE editing}) and pushed every layer-$16$ hidden state through the SAE encoder. For each reported feature, we kept the three word positions across that entire pool with the largest SAE activation, and printed a $\sim$100-character window of generated text around each one. This "top-activating contexts" practice for interpreting individual SAE features follows~\citet{bricken2023towards} and~\citet{templeton2024scaling}.
To better understand Table \ref{tab:feature_examples_L16}, here we explain each of its columns: \texttt{Feature}: the SAE coordinate index $j$.
\texttt{Role/target}: the per-error-type label assigned in \S\ref{method:features} (suppress / boost) with that feature's causal $\Delta_t^{(j)}$ on its target error type, repeated here for context (not re-measured in this table).
\texttt{Act.}: the SAE activation value $z_{j}$ of feature $j$ at that single word position. Concretely it is one dot product plus one scalar bias,
\begin{equation*}
z_{j} \;=\; \bigl(h - b_{\text{dec}}\bigr)^{\top}\, W_{\text{enc}}[:,j] \;+\; b_{\text{enc},j},
\end{equation*}
where $h \in \mathbb{R}^{4096}$ is the layer-$16$ residual-stream hidden state at that word and $W_{\text{enc}}[:,j]$ is feature $j$'s learned "fingerprint" direction (the centred-linear encoder of~\citet{bricken2023towards}; we use the TopK variant of~\citet{gao2025scaling}). Bigger $z_{j}$ means $h$ aligns more strongly with that direction; we bypass the per-token Top-$K$ mask so $z_{j}$ is interpretable even for features outside that token's top-$K$.
\texttt{Generated-report context}: a $\sim$100-character window of the unsteered report around that word, plus the source \texttt{study\_id}.

It is worth mentioning that the Table~\ref{tab:feature_examples_L16} is not a baseline-vs-steered comparison. The SAE only \emph{inspects} activations in this table; it does not edit them; every report shown is unsteered RadVLM output. The diagnostic answers "where in the unsteered output does each feature activate?", not "how much did steering improve the report?". The steering result is in Tables~\ref{tab:main_results}, and \ref{tab:per_error}.

Each feature has a causal sign that tells us what it does. A \emph{boost} feature has $\Delta_t{>}0$: turning it off makes error type $t$ worse, so it was helping the model when it was activated. A \emph{suppress} feature has $\Delta_t{<}0$: turning it off makes the error $t$ better, so it was driving that error when it activated. The text snippets in Table~\ref{tab:feature_examples_L16} therefore have a simple interpretation: boost firings show what the model looks like when it is doing something right, and suppress activations show what it looks like when it is making mistakes.
\paragraph{From snippets to hypothesized patterns.}
Inspecting the top-3 firings of each reported feature, we find that the surrounding text clusters into a small number of recurring phrasings, conventions, and fixed templates that the unsteered model produces repeatedly. For each feature, we summarize our interpretation of the dominant theme as a short label, shown in the Hypothesized pattern column of Table~\ref{tab:feature_examples_L16} as an interpretive annotation aid for the snippets, not a quantity derived from data. 
Across the 12 reported features, the patterns split cleanly along the boost / suppress axis: \emph{boost} features activate on radiology-writing \emph{habits}, and \emph{suppress} features activate on \emph{template traps}. The two paragraphs below describe each class.

\paragraph{Boost features encode radiology-writing habits.}
The boost features activate on three habits of careful radiology writing.
\textbf{Explicit negation}: stating clearly that something is \emph{not} present. \#16965 ($\Delta_{\mathrm{FF}}{=}{+}0.27$) activates on \emph{"There is no evidence of pneumothorax. There is no evidence of pulmonary edema"}.
\textbf{Anatomic checklist}: walking through lungs $\rightarrow$ mediastinum $\rightarrow$ bones. \#5872 ($\Delta_{\mathrm{FF}}{=}{+}0.28$) activates on \emph{"The lungs are clear. The cardiomediastinal silhouette is within normal"}; \#20872 ($\Delta_{\mathrm{MF}}{=}{+}0.23$) continues with \emph{"\ldots the osseous\ldots"}, the cue to enumerate skeletal findings.
\textbf{Listing incidental findings}: noting old fractures, implants, devices. \#1925 ($\Delta_{\mathrm{MF}}{=}{+}0.37$, the largest boost effect) activates on \emph{"Multiple old bilateral rib fractures"} and \emph{"A vagal nerve stimulator is present"}.
Amplifying these features makes the model more thorough and more grounded in the image; turning them off makes the model less careful about them.

\paragraph{Suppress features mark \emph{template traps}.}
A template trap is a fixed phrase the model commits to and produces in full once it begins, regardless of what the image actually shows. Three traps appear among the reported suppress features.

\textbf{End-of-report negation chain.} \#4240 ($\Delta_{\mathrm{WL}}{=}{-}0.09$) activates on \emph{"No pleural effusions. No pulmonary edema. No pneumonia."}.
\textbf{Paired-finding listing.} \#2541 ($\Delta_{\mathrm{MF}}{=}{-}0.09$, $\Delta_{\mathrm{FF}}{=}{+}0.14$) peaks on effusion/atelectasis-style paired clauses in its top contexts; ablation reduces MF on the screening slice despite that wording, so we treat the causal $\Delta$ as authoritative and the snippet as localization only (not a literal map from zeroing the feature to deleting every effusion phrase).
\textbf{Generic-normal severity template.} \#29962 ($\Delta_{\mathrm{WS}}{=}{-}0.11$) activates on \emph{"\ldots The aorta is tortuous. There is no pneumothorax or pleural effusion"}.
\textbf{Confident support-device language.} \#31611 ($\Delta_{\mathrm{FF}}{=}{-}0.16$) activates on \emph{"lead tip in the distal SVC"} and \emph{"right-sided Swan-Ganz catheter"}.
Turning off such a feature does not stop the model from describing real devices or real negative findings; it makes the model less likely to enter the template \emph{by default} when the image does not warrant it.
\paragraph{Opposing roles on the same behaviour.}
Study \texttt{s58637592} appears in the top activation contexts of two features with opposite roles. \#15810 (boost, $\Delta_{\mathrm{FF}}{=}{+}0.30$) activates on \emph{"\ldots is a prosthetic aortic valve. There is a prosthetic aortic valve.\,\ldots"}, the start of a repetition loop; keeping it active prevents the loop from cascading into more false findings. \#8368 (suppress, $\Delta_{\mathrm{WS}}{=}{-}0.10$) activates on the same loop in the same study, but here it contributes to wrong-severity drift. The two features react to the same underlying behaviour and pull in opposite directions on different error types. This is why the steering recipe needs a separate suppress and boost list \emph{per error type} (\S\ref{method:features}): a single global label per feature would get one of the two effects wrong (\ Table~\ref{tab:feature_pertype_summary_L16}).

\begin{table*}[!htbp]
\centering
\scriptsize
\setlength{\tabcolsep}{3pt}
\renewcommand{\arraystretch}{0.9}
\resizebox{0.99\textwidth}{!}{%
\begin{tabular}{@{}ll p{0.16\textwidth} r p{0.42\textwidth}@{}}
\toprule
Feature & Role / target ($\Delta$) & Hypothesised pattern & Act. & Generated-report context (\textellipsis~window around the most-active position) \\
\midrule
\#31611 & \textbf{supp} / $\Delta_{\mathrm{FF}} = -0.16$ & \emph{support-device language} & 10.39 & \ldots{}ead tip in the distal SVC. Heart size is upper limits of normal. There is mild prominence of the pul\ldots{}\hfill \\

 &  &  & 10.21 & \ldots{}e with distal lead tip in the proximal right atrium. There is a right-sided Swan-Ganz catheter with\ldots{}\hfill \\
 &  &  & 10.18 & \ldots{}e with the distal lead tip in the distal SVC. There is an enteric tube whose side port is at the GE\ldots{}\hfill \\
\addlinespace[2pt]
\#15810 & \textbf{boost} / $\Delta_{\mathrm{FF}} = +0.30$ & \emph{repetition-loop brake} & 18.42 & \ldots{}is a prosthetic aortic valve. There is a prosthetic aortic valve. There\hfill \\
 &  &  & 18.31 & \ldots{}is a prosthetic aortic valve. There is a prosthetic aortic valve. There is a prosthetic aortic valv\ldots{}\hfill \\
 &  &  & 18.28 & \ldots{}is a prosthetic aortic valve. There is a prosthetic aortic valve. There is a prosthetic aortic valv\ldots{}\hfill \\
\addlinespace[2pt]
\#2541 & \textbf{supp} / $\Delta_{\mathrm{MF}} = -0.09$ & \emph{paired-finding listing} & 16.86 & \ldots{}e left pleural effusion with associated left basilar atelectasis. There is no pneumothorax. The righ\ldots{}\hfill \\
 &  &  & 16.81 & \ldots{}ht greater than left. There is pulmonary vascular congestion and mild pulmonary edema. The cardiac s\ldots{}\hfill \\
 &  &  & 16.80 & \ldots{}e left pleural effusion with associated left basilar atelectasis. The right lung is clear. There is\ldots{}\hfill \\
\addlinespace[2pt]
\#1925 & \textbf{boost} / $\Delta_{\mathrm{MF}} = +0.37$ & \emph{incidental findings} & 10.02 & \ldots{}reflect areas of atelectasis. No pneumothorax is identified. Multiple old bilateral rib fractures a\ldots{}\hfill \\
 &  &  & 9.96 & \ldots{}A vagal nerve stimulator is present. Multiple old right rib fractures are noted.\hfill \\
 &  &  & 9.81 & \ldots{}al. A vagal nerve stimulator is present. Multiple old right rib fractures are noted.\hfill \\
\addlinespace[2pt]
\#4240 & \textbf{supp} / $\Delta_{\mathrm{WL}} = -0.09$ & \emph{negation-chain template} & 9.52 & \ldots{}tion. Moderate cardiomegaly. No pleural effusions. No pulmonary edema. No pneumonia.\hfill \\
 &  &  & 9.40 & \ldots{}rdiomegaly. No pleural effusions. No pulmonary edema. No pneumonia.\hfill \\
 &  &  & 9.36 & \ldots{}ax. No pleural effusions. No pulmonary edema.\hfill \\
\addlinespace[2pt]
\#28456 & \textbf{boost} / $\Delta_{\mathrm{WL}} = +0.06$ & \emph{anatomic detail} & 7.64 & \ldots{}umothorax. There are low lung volumes. Bibasilar opacities are a combination of atelectasis and smal\ldots{}\hfill \\
 &  &  & 6.96 & \ldots{}x. There are low lung volumes. Bibasilar opacities are a combination of atelectasis and small effusi\ldots{}\hfill \\
 &  &  & 6.84 & \ldots{}C. There is no pneumothorax. Bibasilar opacities are a combination of small effusions and adjacent a\ldots{}\hfill \\
\addlinespace[2pt]
\#29962 & \textbf{supp} / $\Delta_{\mathrm{WS}} = -0.11$ & \emph{generic-normal template} & 7.94 & \ldots{}is normal. The aorta is tortuous. There is no pneumothorax or pleural effusion. There is no focal c\ldots{}\hfill \\
 &  &  & 7.87 & \ldots{}ildly enlarged. The aorta is tortuous. There is no pleural effusion or pneumothorax. The lungs appea\ldots{}\hfill \\
 &  &  & 7.83 & \ldots{}ildly enlarged. The aorta is tortuous. There is no pleural effusion or pneumothorax. The lungs appea\ldots{}\hfill \\
\addlinespace[2pt]
\#7046 & \textbf{boost} / $\Delta_{\mathrm{WS}} = +0.05$ & \emph{precise laterality} & 13.62 & \ldots{}ve subcutaneous emphysema in the right neck and chest wall. There is no pneumothorax. There are mode\ldots{}\hfill \\
 &  &  & 13.61 & \ldots{}f subcutaneous emphysema in the right chest wall. There is a small amount of atelectasis at the righ\ldots{}\hfill \\
 &  &  & 13.58 & \ldots{}omediastinal silhouette is within normal limits. Right chest wall port is seen with catheter tip in\ldots{}\hfill \\
\addlinespace[2pt]
\#5872 & \textbf{boost} / $\Delta_{\mathrm{FF}} = +0.28$ & \emph{anatomic checklist} & 20.83 & The lungs are clear. The cardiomediastinal silhouette is within normal\ldots{}\hfill \\
 &  &  & 20.72 & The lungs are clear. The cardiomediastinal silhouette is within normal\ldots{}\hfill \\
 &  &  & 20.71 & The lungs are clear. The cardiomediastinal silhouette is within normal\ldots{}\hfill \\
\addlinespace[2pt]
\#16965 & \textbf{boost} / $\Delta_{\mathrm{FF}} = +0.27$ & \emph{explicit negation} & 7.83 & \ldots{}idence of pneumothorax. The lung volumes are normal. Normal size of the cardiac silhouette. Normal h\ldots{}\hfill\\
 &  &  & 7.56 & There is no evidence of pneumothorax. There is a small left pleural effusion. There is no\ldots{}\hfill \\
 &  &  & 7.54 & There is no evidence of pneumothorax. There is no evidence of pulmonary edema. There is no\ldots{}\hfill \\

\addlinespace[2pt]
\#20872 & \textbf{boost} / $\Delta_{\mathrm{MF}} = +0.23$ & \emph{anatomic checklist (osseous)} & 10.43 & The lungs are clear. There is no pneumothorax or pleural effusion. The osseous\ldots{}\hfill \\

 &  &  & 8.92 & The lungs are clear. The cardiomediastinal silhouette\ldots{}\hfill \\

 &  &  & 8.55 & The lungs are clear. The cardiomediastinal silhouette is within normal\ldots{}\hfill \\

\addlinespace[2pt]
\#8368 & \textbf{supp} / $\Delta_{\mathrm{WS}} = -0.10$ & \emph{Late-report repetition / devices
} & 10.28 & \ldots{}s a prosthetic pulmonary valve. There is a prosthetic pulmonic valve. There is a prosthetic pulmonic\ldots{}\hfill  \\

 &  &  & 9.92 & \ldots{}There is a nasogastric tube whose distal tip and side port are below the GE junction. There is a lef\ldots{}\hfill  \\

 &  &  & 9.67 & \ldots{}here is mild pulmonary edema. There is atelectasis at the lung bases.\hfill  \\

\addlinespace[2pt]
\bottomrule
\end{tabular}%
}
\caption{\textbf{Generated-report contexts where reported SAE features respond most strongly (RadVLM, layer~16, MIMIC-CXR validation set).} For each reported feature from Table~\ref{tab:feature_pertype_summary_L16} and the per-layer top features (Table~\ref{tab:feature_pertype_top_L16}).}
\label{tab:feature_examples_L16}
\end{table*}

\paragraph{Synthesis and link to the cross-model functional overlap.}
Previously, we observed that: \emph{boost features amplify radiology-writing habits, suppress features close the model's own template traps, and combined steering applies both.}
This interpretation also explains the asymmetry observed in the cross-model census (\S\ref{sec:res_census}, Table~\ref{tab:census}).
Habits such as explicit negation and anatomic checklist are general radiology conventions that any CXR-trained VLM is likely to encode, and indeed the boost directions transfer across the three architectures we tested (Jaccard $0.65$--$0.75$, cosine $\!>\!0.91$).
Template traps are different: each model has its own, because the templates a model defaults into are shaped by its own training rather than by general radiology conventions. RadVLM's particular end-of-report negation chain and its repetition loop on prosthetic-valve descriptions are not shared by LLaVA-Rad or CheXOne, and the suppress directions therefore do not transfer (Jaccard $0.41$--$0.58$).
The steering \emph{recipe} is the same in every model; only the suppress \emph{list} must be re-derived per model.
Aggregated over the test set (Table~\ref{tab:per_error}), the steered model recovers more findings the reference contains ($-654$ missed-finding errors) but also adds findings the reference does not ($+1,112$ false-finding errors), exactly what we expect from a recipe that pushes the model toward more thorough reporting (boost) while removing template fragments that were sometimes real and sometimes fabricated (suppress).

\paragraph{Quantitative analysis of activation features.}
The qualitative patterns above, as boost features on radiology habits, suppress features on template traps, bring another question: \emph{where in the generated report does each feature actually activate?}. We answer this by using the generated reports from unsteered RadVLM on 10,000 validation studies, and activated features (Activation > 2.0) on every token at layer 16. For each reported feature, Table~\ref{feature_profile_L16} shows how many studies it activates on, how strongly it activates on average, and at which point in the report it activates most (early, middle, or late). By analyzing this, we observed three patterns as follows:
\begin{itemize}[leftmargin=*,nosep]
\item \textbf{Features related to the repetition-loop are rare and late.} Feature \#15810 as FF-boost activates on only 43 out of 9,951 studies, but when it activates, 87\% of it occurs in the second half of the report. And all three of its top activation snippets in Table~\ref{tab:feature_examples_L16} are literal phrase replications. It is not a clinically related observation; it is a specific alarm for a particular failure mode that happens in generation during late steps. 
\item \textbf{"Good Reporting" features activate everywhere.} Features like \#5872 and \#1925 (both FF-boost) fire on every single validation study. \#5872 has the highest average activation ($\bar{a}{=}8.43$). These are the broad "write a proper radiology report" features that our boost-side amplification supports.
\item \textbf{Template-trap features concentrate at the end.}
Feature \#4240 (WL-suppress) fires $98\%$ of the time in the second half of the report, with a median position of $0.92$ (nearly the last token).
This matches the canonical end-of-report negation list ("\emph{No pleural effusions. No pulmonary edema. No pneumonia.}") as a fixed phrase the model appends by default, regardless of image content.
Zeroing it produces more wrong-location errors ($\Delta_{\mathrm{WL}}=-0.09$), which is exactly what we would expect if the feature controls an often-useful but sometimes-incorrect template.

\end{itemize}
\begin{table}[h]
\centering
\footnotesize
\setlength{\tabcolsep}{2pt}
\resizebox{\columnwidth}{!}{%
\begin{tabular}{@{}llrrrrrr@{}}
\toprule
Feature & Role/$\Delta$ & \#Active & \#Stud. & $\bar{a}$ & $p_{50}^{\text{pos}}$ & frac.\,late & rep. \\
\midrule
\#31611 & \textbf{supp}/$\mathrm{FF}\,-0.16$ & 34,058 & 2,853 & 3.89 & 0.61 & 65\% & 1/3 \\
\#15810 & \textbf{boost}/$\mathrm{FF}\,+0.30$ & 932 & 43 & 5.67 & 0.76 & 87\% & 3/3 \\
\#2541 & \textbf{supp}/$\mathrm{MF}\,-0.09$ & 20,972 & 6,798 & 7.12 & 0.63 & 66\% & 0/3 \\
\#1925 & \textbf{boost}/$\mathrm{MF}\,+0.37$ & 167,720 & 9,951 & 3.22 & 0.43 & 44\% & 0/3 \\
\#4240 & \textbf{supp}/$\mathrm{WL}\,-0.09$ & 2,203 & 825 & 4.08 & 0.92 & 98\% & 0/3 \\
\#28456 & \textbf{boost}/$\mathrm{WL}\,+0.06$ & 47,526 & 4,952 & 2.78 & 0.49 & 50\% & 0/3 \\
\#29962 & \textbf{supp}/$\mathrm{WS}\,-0.11$ & 2,321 & 390 & 2.85 & 0.65 & 76\% & 1/3 \\
\#7046 & \textbf{boost}/$\mathrm{WS}\,+0.05$ & 917 & 502 & 6.28 & 0.63 & 69\% & 0/3 \\
\#5872 & \textbf{boost}/$\mathrm{FF}\,+0.28$ & 43,626 & 9,951 & 8.43 & 0.66 & 65\% & 0/3 \\
\#16965 & \textbf{boost}/$\mathrm{FF}\,+0.27$ & 863 & 290 & 4.25 & 0.61 & 70\% & 1/3 \\
\#20872 & \textbf{boost}/$\mathrm{MF}\,+0.23$ & 440 & 152 & 3.05 & 0.17 & 25\% & 0/3 \\
\#8368 & \textbf{supp}/$\mathrm{WS}\,-0.10$ & 89,383 & 7,294 & 3.47 & 0.76 & 87\% & 1/3 \\
\bottomrule
\end{tabular}%
}
\caption{\textbf{Activation profile of reported features at layer-16  on the RadVLM per-token validation pool} ($10{,}000$ decodes; distinct \texttt{study\_id} counts in \#Stud.; pre-activation threshold $> 2.00$; causal $\Delta$ uses $N{=}100$ studies per feature, \S\ref{method:features}). \#Active = number of token positions above threshold; \#Stud.\ = distinct study ids on which the feature is active; $\bar{a}$ = mean pre-activation when active; $p_{50}^{\text{pos}}$ = median normalised decode position ($0$ = first token of the report, $1$ = last); frac.\,late = fraction of activations in the second half of the report; rep.\ = number of top-3 activation-context snippets (Table~\ref{tab:feature_examples_L16}) that contain a literal phrase repetition.}
\label{feature_profile_L16}
\end{table}

\noindent Figure~\ref{fig:feature_profile_hist_L16} shows the full decode-position distributions for all 12 reported features.
In short, each feature activates at the part of the report where its effect matters: boost features that prevent repetition are most active toward the end, where repetition occurs; suppress features tied to standard phrasing are active wherever that phrasing appears. The position patterns are consistent with the causal roles, providing evidence that the SAE has learned an interpretable structure rather than arbitrary directions.

\section{Cross-model feature analysis: formal definition}
\label{sec:appendix_census_definition}

This section gives the exact recipe behind the cross-model census reported in Table~\ref{tab:census}.
.

\paragraph{Setup.}
Let~$\small \mathcal{T}=\{\mathrm{FF},\mathrm{MF},\mathrm{WL},\mathrm{WS}\}$ denote the four GREEN error types we steer.
For each model $m\in\{\textsc{RadVLM},\textsc{LLaVA-Rad},\textsc{CheXOne}\}$, layer $\ell\in\{8,16,20,24\}$ and direction $d\in\{\textsc{suppress},\textsc{boost}\}$, the per-error-type causal screen of \S\ref{method:features} (run with \texttt{green\_pertype} scoring) returns, for each candidate feature $f$, the 4-vector
\begin{equation}\small
\Delta_f^{(m,\ell)} \;=\; \bigl(\Delta_{\mathrm{FF}},\Delta_{\mathrm{MF}},\Delta_{\mathrm{WL}},\Delta_{\mathrm{WS}}\bigr)_f
\;\in\;\mathbb{R}^4
\label{eq:per_feature_delta}
\end{equation}
where $\Delta_{t,f}=\mathbb{E}_{x}\bigl[\#\,\mathrm{err}_t(y^{\text{abl}_f}(x))-\#\,\mathrm{err}_t(y^{\text{base}}(x))\bigr]$ is the mean change in $t$-type error count when feature $f$ is zeroed at layer $\ell$, averaged over the validation samples that passed the word-F1 prefilter.
The causal-screen output also ships pre-ranked, sign-consistent target lists $T_t^{(m,\ell,d)}=$ \texttt{intervention\_targets[$d$][$t$]}, ordered by $|\Delta_{t,f}|$ with sign matched to $d$ (suppress: $\Delta_t<0$; boost: $\Delta_t>0$).

\paragraph{Consensus set.}
For each $(m,\ell,d)$ we form a deterministic consensus list $C^{(m,\ell,d)}$ of size $N=100$ by round-robin-merging the four ranked per-type lists, taking the first occurrence of each feature:
\begin{equation}\scriptsize
C^{(m,\ell,d)} \;=\; \mathrm{RR}\bigl(T_{\mathrm{FF}}^{(m,\ell,d)},\,T_{\mathrm{MF}}^{(m,\ell,d)},\,T_{\mathrm{WL}}^{(m,\ell,d)},\,T_{\mathrm{WS}}^{(m,\ell,d)}\bigr)_{1:N}
\label{eq:consensus_set}
\end{equation}
This gives every error type proportional representation while preserving the per-type rank order.
Sizes $n_a, n_b$ in Table~\ref{tab:census} report $|C^{(a,\ell,d)}|,|C^{(b,\ell,d)}|$ (smaller than $N$ when fewer than $N$ features pass the screen, e.g.\ a thinly-populated boost direction on \textsc{CheXOne} at $\ell=8$).

\paragraph{Two summaries per (model, layer, direction).}
Each model's behaviour at $(\ell,d)$ is summarised by two 4-vectors derived from $C^{(m,\ell,d)}$.
The \emph{signature vector} is the mean per-feature causal effect over the consensus set:
\begin{equation}
\mathbf{s}^{(m,\ell,d)} \;=\; \frac{1}{\bigl|C^{(m,\ell,d)}\bigr|}\sum_{f\in C^{(m,\ell,d)}} \Delta_f^{(m,\ell)} \;\in\;\mathbb{R}^4.
\label{eq:signature}
\end{equation}
The \emph{categorical profile} is the empirical distribution of dominant, direction-consistent error types across the consensus set:
\begin{equation}\scriptsize
\pi^{(m,\ell,d)}_{t} \;=\; \frac{1}{V}\sum_{f\in C^{(m,\ell,d)}}\!\mathbf{1}\!\left[\,t \;=\;\!\!\!\!\arg\max_{t'\in\mathcal{T}\,:\,\mathrm{sgn}(\Delta_{t',f})=\mathrm{sgn}(d)}\!\!\!\!|\Delta_{t',f}|\,\right]
\label{eq:profile}
\end{equation}
where $\small V=\sum_{f}\mathbf{1}[\exists\,t'\!:\,\mathrm{sgn}(\Delta_{t',f})=\mathrm{sgn}(d)]$ is the number of features that have at least one direction-consistent component, and $\mathrm{sgn}(\textsc{suppress})=-1$, $\mathrm{sgn}(\textsc{boost})=+1$

\paragraph{Pairwise similarities.}
For an unordered model pair $(a,b)$ and a given $(\ell,d)$, the two reported numbers in Table~\ref{tab:census} are
\begin{equation}\small
\mathrm{signature\_cosine}_{a,b} \;=\; \frac{\mathbf{s}^{(a)}\cdot\mathbf{s}^{(b)}}{\bigl\lVert\mathbf{s}^{(a)}\bigr\rVert_2\,\bigl\lVert\mathbf{s}^{(b)}\bigr\rVert_2}\;\in\;[-1,\,1]
\label{eq:signature_cosine}
\end{equation}
and the Ruzicka (weighted-Jaccard) similarity of the categorical profiles,
\begin{equation}\small
\mathrm{functional\_jaccard}_{a,b} \;=\; \frac{\sum_{t\in\mathcal{T}}\min\!\bigl(\pi^{(a)}_{t},\pi^{(b)}_{t}\bigr)}{\sum_{t\in\mathcal{T}}\max\!\bigl(\pi^{(a)}_{t},\pi^{(b)}_{t}\bigr)}\;\in\;[0,\,1]
\label{eq:functional_jaccard}
\end{equation}
Cosine in Eq.~\ref{eq:signature_cosine} compares the \emph{direction and magnitude profile} of the two models' causal effects in $\mathcal{T}$-space; Ruzicka in Eq.~\ref{eq:functional_jaccard} compares the discrete \emph{which-error-type-dominates} distribution.
Both are basis-free and therefore well defined across architectures that do not share SAE indices.

\paragraph{Aggregation and CIs.}
Table~\ref{tab:census} reports, per direction and per model pair, the mean of \eqref{eq:signature_cosine} and \eqref{eq:functional_jaccard} over the four steered layers $\ell\in\{8,16,20,24\}$.
The bracketed $95\%$ confidence intervals are percentile bootstrap CIs over those four layers ($B=10^4$ resamples).
Boost-vs-suppress contrasts use a separate paired bootstrap on the difference of layer means.

\section{Cross-dataset transfer: IU-Xray}
\label{sec:appendix_iuxray}

The SAEs and identifies features in the paper are based on MIMIC-CXR. To test whether the steering recipe transfers across datasets,  not just across model architectures, we evaluate the \emph{same} SAE weights, feature lists, and operating point ($\alpha=0.20$, $K=20$, layers $\{8,16,20,24\}$, Combined mode) on \textbf{IU-Xray}, a publicly available collection of $3,955$ frontal chest radiographs with paired radiology reports from Indiana University.
After filtering for image availability, $N=3,307$ studies are decoded with unsteered and SAE-steered RadVLM.

Table~\ref{tab:iuxray_headline} shows the results of the steered-RadVLM on IU-Xray. We also report the per-error-type breakdown on IU-Xray. The dominant pattern matches MIMIC-CXR (Table~\ref{tab:per_error}): missing findings drop sharply ($-850$) while false findings increase moderately ($+223$).
  Unlike MIMIC-CXR, total significant errors \emph{decrease} on IU-Xray ($-713$), because the larger MF reduction on shorter reports more than offsets the FF increase.

\begin{table}[h]
  \centering
  \footnotesize
  \setlength{\tabcolsep}{3.5pt}
  \renewcommand{\arraystretch}{1.05}
  \resizebox{\columnwidth}{!}{%
  \begin{tabular}{@{}lcccccccc@{}}
    \toprule
    \textbf{Setting} & \textbf{GREEN}$\uparrow$ & \textbf{RG}$\uparrow$ & \textbf{CXB$_{\mu}$}$\uparrow$ & \textbf{BS}$\uparrow$ & \textbf{BL}$\uparrow$ & \textbf{RL}$\uparrow$ & \textbf{Comp.}$\uparrow$ \\
    \midrule
    \multicolumn{8}{@{}l@{}}{\textbf{RadVLM} (Qwen3-VL 8B) on IU-Xray --- $\alpha{=}0.20$, $K{=}20$} \\
    \quad Base            & 49.7 & 24.7 & 49.2 & 88.6 & 5.6 & 26.6 & 46.5 \\
    \quad + SAE Combined  & \textbf{53.6} & \textbf{28.9} & 47.0 & \textbf{88.7} & \textbf{6.5} & \textbf{26.7} & \textbf{49.0} \\
    \quad \textit{$\Delta$ (pp)} & \textit{+3.8} & \textit{+4.1} & \textit{$-2.2$} & \textit{+0.2} & \textit{+0.9} & \textit{+0.1} & \textit{+2.5} \\
    \quad \textit{rel.\ \%}     & \textit{+7.7\%} & \textit{+16.7\%} & \textit{$-4.5\%$} & \textit{+0.2\%} & \textit{+15.7\%} & \textit{+0.3\%} & \textit{+5.4\%} \\
    \bottomrule
  \end{tabular}%
  }
  \caption{RadVLM headline results on \textbf{IU-Xray} ($N=3,307$ reports), using the same SAE weights, feature lists, and operating point ($\alpha=0.20$, $K=20$, layers $\{8,16,20,24\}$) as the MIMIC-CXR test in Table~\ref{tab:main_results}.
  GREEN, RadGraph, BERTScore, BLEU-4, and ROUGE-L all improve; CheXbert 14-label micro F1 drops slightly ($-4.5\%$~rel.), but the net Composite gain is $+2.5$\,pp ($+5.4\%$~rel.).
  Column definitions match Table~\ref{tab:main_results}.}
  \label{tab:iuxray_headline}
\end{table}

\begin{table}[h]
  \centering
  \footnotesize
  \setlength{\tabcolsep}{4pt}
  \resizebox{\columnwidth}{!}{%
  \begin{tabular}{@{}lrrr@{}}
    \toprule
    \textbf{GREEN error type} & \textbf{Baseline} & \textbf{+ SAE Combined} & \textbf{$\Delta$} \\
    \midrule
    False finding (FF) & 2,140 & 2,363 & $+223$ \\
    Missing finding (MF) & 6,498 & 5,648 & $\mathbf{-850}$ \\
    Wrong location (WL) & 203 & 279 & $+76$ \\
    Wrong severity (WS) & 310 & 245 & $-65$ \\
    False comparison (FC) & 299 & 221 & $-78$ \\
    Missing comparison (MC) & 115 & 96 & $-19$ \\
    \midrule
    \textbf{Total significant errors} & \textbf{9,565} & \textbf{8,852} & $\mathbf{-713}$ \\
    \bottomrule
  \end{tabular}%
  }
  \caption{Per-error-type breakdown on \textbf{IU-Xray}, at the same operating point as Table~\ref{tab:iuxray_headline}.
  The dominant pattern matches MIMIC-CXR (Table~\ref{tab:per_error}): missing findings drop sharply ($-850$) while false findings increase moderately ($+223$).
  Unlike MIMIC-CXR, total significant errors \emph{decrease} on IU-Xray ($-713$), because the larger MF reduction on shorter reports more than offsets the FF increase.}
  \label{tab:iuxray_pererror}
\end{table}

\section{Quantitative examples: Improvement, regression}
Table~\ref{tab:iuxray_qual_examples} shows four IU-Xray cases annotated with GREEN error tags derived directly from GREEN model raw completions.
IU-Xray is a public dataset, so we reproduce the original chest radiographs and full report text.
The steering removes a false cardiomegaly report on an otherwise normal study (iuxray\_002410, $GR{:}\,0.67{\to}1.00$); recovers a hemodialysis catheter and its tip position that the baseline omitted entirely (iuxray\_000286, $GR{:}\,0.17{\to}0.80$); and nudges a borderline-enlarged heart from "within normal limits" to "top normal in size",  though the RLL granuloma remains unreported (iuxray\_001563, $GR{:}\,0.00{\to}0.75$).
The regression case (iuxray\_001044) shows the FF$\uparrow$ trade-off: the baseline accurately describes a subclavian catheter and a normal study ($GR{=}1.00$); steering hallucinates a pleural effusion and atelectasis ($GR{:}\,1.00{\to}0.20$).

\begin{table*}[t]
\centering
\scriptsize
\setlength{\tabcolsep}{3pt}
\renewcommand{\arraystretch}{0.90}
\setlength{\AppendixQualRowHeight}{0.135\textheight}
\begin{tabular}{@{}c p{0.11\textwidth} p{0.17\textwidth} p{0.17\textwidth} p{0.17\textwidth} c@{}}
\toprule
\textbf{Case} & \textbf{CXR} & \textbf{Ground truth} & \textbf{Baseline (unsteered)} & \textbf{+SAE steering} & \shortstack{\textbf{GREEN}\\[-2pt]\scriptsize(bl.$\to$st.)} \\
\midrule
%% --- Improvement 1 (iuxray_002410: false cardiomegaly removed → GREEN 1.00) ---
\AppendixQualCaseVert{Improvement}
&
\parbox[c][\AppendixQualRowHeight]{0.12\textwidth}{%
  \centering
  \includegraphics[width=\linewidth,height=0.13\textheight,keepaspectratio]{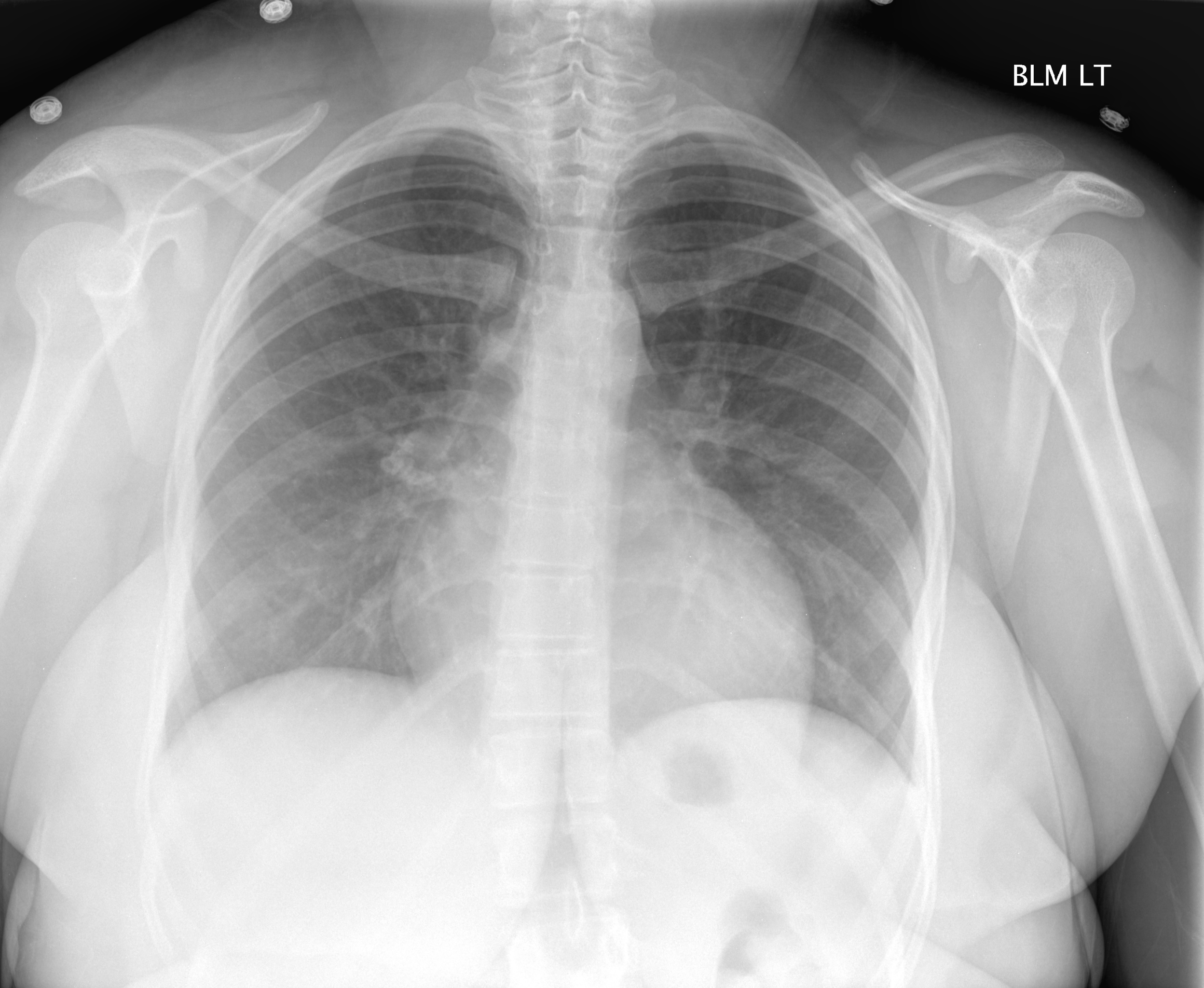}\\[1pt]
  {\tiny\sffamily iuxray\_002410}\\[1pt]
  {\tiny\sffamily\color{gray!55!black}normal study}}
&
The cardiomediastinal silhouette is within normal limits for size and contour.\ The lungs are normally inflated without evidence of focal airspace disease, pleural effusion, or pneumothorax.\ Osseous structures are within normal limits for patient age.
&
\textcolor{teal}{The lungs are clear without focal consolidation}\ApqGok{OK}.\ \textcolor{teal}{No pleural effusion or pneumothorax is seen}\ApqGok{OK}.\ \textcolor{red}{The cardiac silhouette is mildly enlarged}\ApqGerr{FF}\@.\ No pulmonary edema is seen\ApqGinsig{INSIG}\@.
&
\textcolor{teal}{The lungs are clear without focal consolidation, pleural effusion, or pneumothorax}\ApqGok{OK}.\ \textcolor{teal}{The cardiomediastinal silhouette is within normal limits}\ApqGok{OK}.\ \textcolor{teal}{The visualized osseous structures are unremarkable}\ApqGok{OK}.
& {\footnotesize$0.67\!\to\!1.00$}
\\[2pt]
\noalign{\vskip 1pt\hrule height 0.3pt\vskip 3pt}
%% --- Improvement 2 (iuxray_000286: hemodialysis catheter recovered → GREEN 0.80) ---
\AppendixQualCaseVert{Improvement}
&
\parbox[c][\AppendixQualRowHeight]{0.12\textwidth}{%
  \centering
  \includegraphics[width=\linewidth,height=0.13\textheight,keepaspectratio]{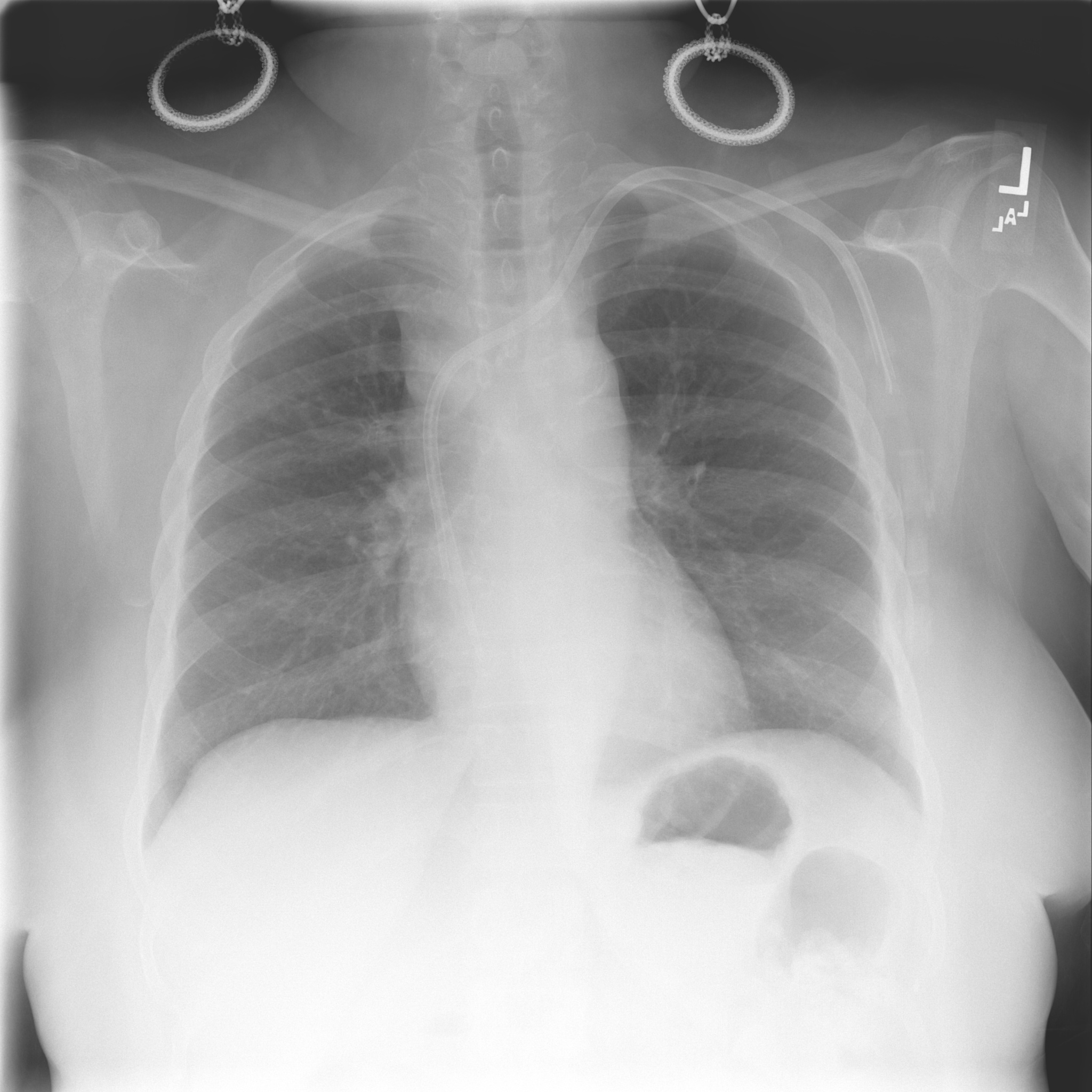}\\[1pt]
  {\tiny\sffamily iuxray\_000286}\\[1pt]
  {\tiny\sffamily\color{gray!55!black}hemodialysis catheter}}
&
A left-sided hemodialysis catheter is in place with its distal tip at the right atrium.\ The cardiac silhouette and mediastinal contours are within normal limits.\ There is no focal opacity.\ There is no pneumothorax.\ No large pleural effusion.
&
\textcolor{red}{The lungs are clear}\ApqGerr{FF}\@.\ \textcolor{teal}{The cardiomediastinal silhouette is within normal limits}\ApqGok{OK}.\ No acute osseous abnormalities.
&
\textcolor{orange}{There is a left-sided central venous catheter with the tip in the right atrium}\ApqGwarn{MF}.\ \textcolor{teal}{The lungs are clear without focal consolidation, pleural effusion, or pneumothorax}\ApqGok{OK}.\ \textcolor{teal}{The cardiomediastinal silhouette is normal}\ApqGok{OK}.
& {\footnotesize$0.17\!\to\!0.80$}
\\[2pt]
\noalign{\vskip 1pt\hrule height 0.3pt\vskip 3pt}
%% --- Improvement 3 (iuxray_001563: borderline heart + granuloma → GREEN 0.75) ---
\AppendixQualCaseVert{Improvement}
&
\parbox[c][\AppendixQualRowHeight]{0.12\textwidth}{%
  \centering
  \includegraphics[width=\linewidth,height=0.13\textheight,keepaspectratio]{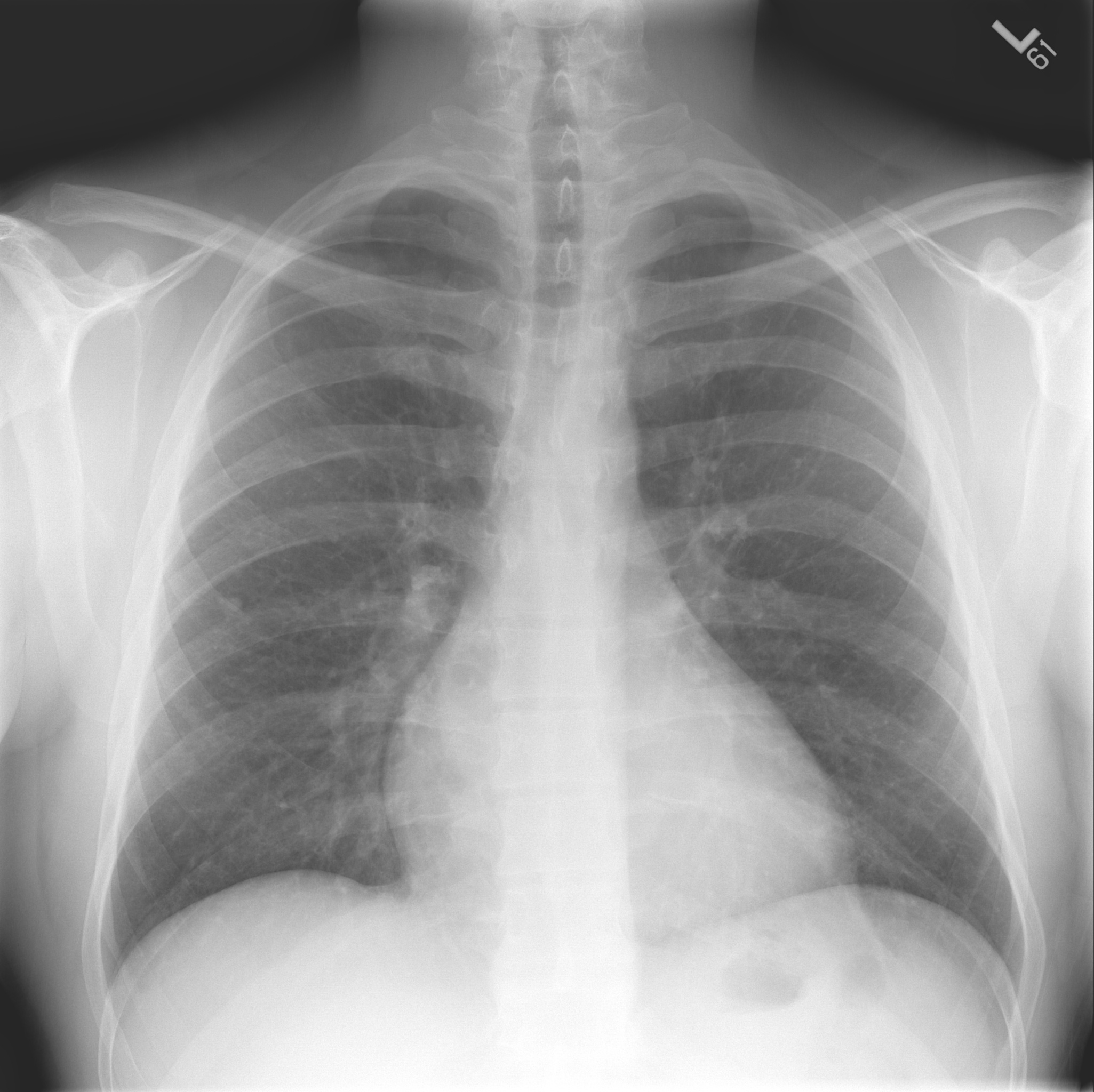}\\[1pt]
  {\tiny\sffamily iuxray\_001563}\\[1pt]
  {\tiny\sffamily\color{gray!55!black}borderline heart,}\\[-1pt]
  {\tiny\sffamily\color{gray!55!black}RLL granuloma}}
&
Heart size borderline enlarged.\ No focal alveolar consolidation, no definite pleural effusion seen.\ No typical findings of pulmonary edema.\ Dense nodule in the right lower lobe suggests a previous granulomatous process.
&
The lungs are clear.\ There is no pneumothorax.\ \textcolor{red}{The heart and mediastinum are within normal limits}\ApqGerr{FF}\@.\ Regional bones and soft tissues are unremarkable.
&
\textcolor{teal}{The lungs are clear without focal consolidation, pleural effusion or pneumothorax}\ApqGok{OK}.\ The pulmonary vasculature is not engorged.\ The cardiac silhouette is top normal in size\ApqGinsig{WS}.\ The mediastinal and hilar contours are within normal limits.\ No acute osseous abnormality is detected.
& {\footnotesize$0.00\!\to\!0.75$}
\\[2pt]
\noalign{\vskip 1pt\hrule height 0.3pt\vskip 3pt}
%% --- Regression (iuxray_000779: moderate FF on enlarged heart study) ---
\AppendixQualCaseVert{Regression}
&
\parbox[c][\AppendixQualRowHeight]{0.12\textwidth}{%
  \centering
  \includegraphics[width=\linewidth,height=0.13\textheight,keepaspectratio]{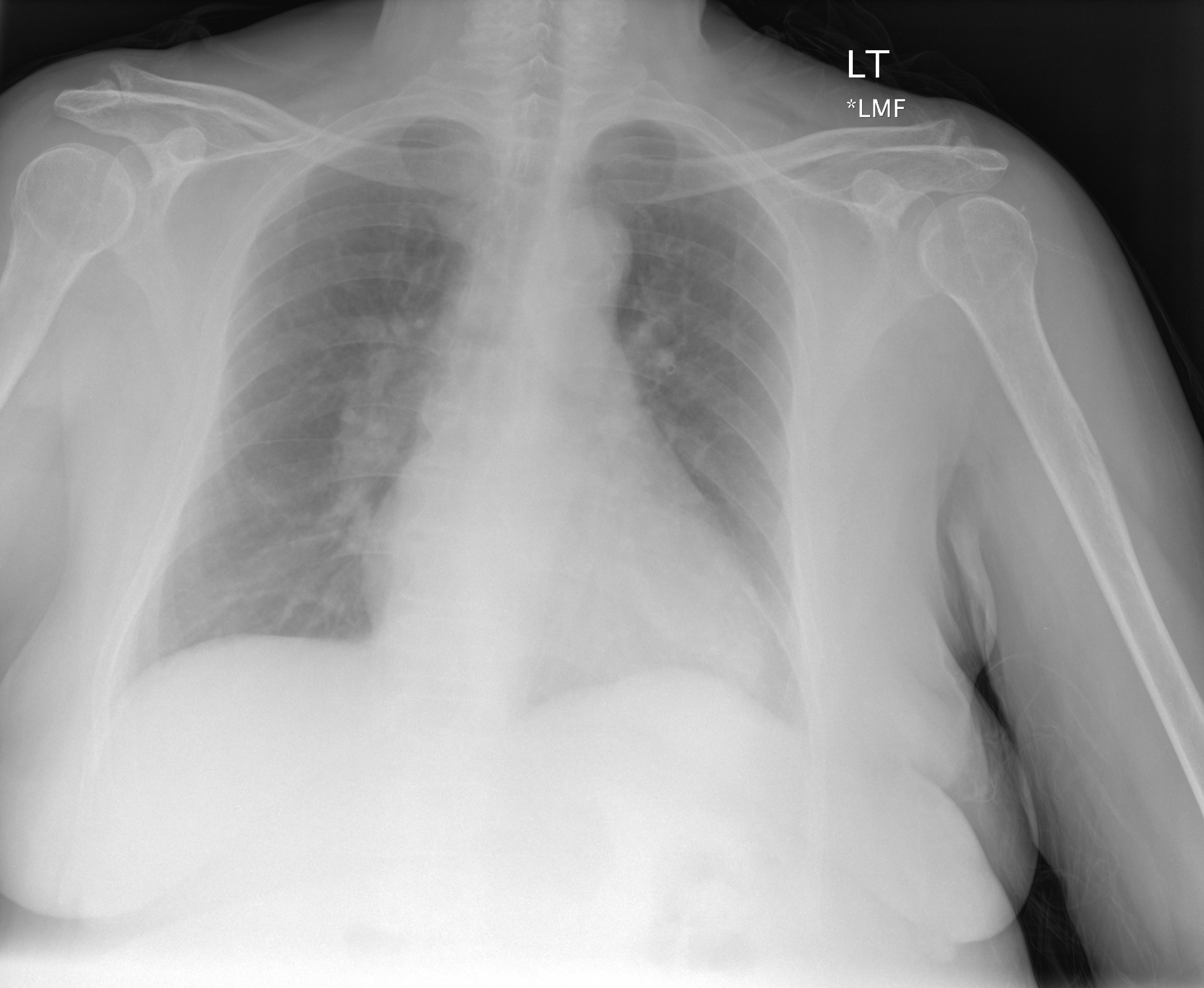}\\[1pt]
  {\tiny\sffamily iuxray\_000779}\\[1pt]
  {\tiny\sffamily\color{gray!55!black}slightly large heart}}
&
The heart is slightly large.\ Pulmonary vasculature is normal.\ No infiltrates.
&
\textcolor{teal}{The lungs are clear without focal consolidation}\ApqGok{OK}.\ No pleural effusion or pneumothorax is seen.\ \textcolor{teal}{The cardiac silhouette is mildly enlarged}\ApqGok{OK}.\ The aorta is calcified and tortuous.
&
The lungs are clear without focal consolidation, pleural effusion, or pneumothorax.\ \textcolor{teal}{The cardiac silhouette is enlarged}\ApqGok{OK}.\ The mediastinal and hilar contours are within normal limits.\ \textcolor{red}{There is a mild compression deformity of a lower thoracic vertebral body}\ApqGerr{FF}\@.
& {\footnotesize$1.00\!\to\!0.50$}
\\
\bottomrule
\end{tabular}
\caption{\textbf{IU-Xray qualitative steering cases} (RadVLM, Combined boost$+$suppress, $\alpha{=}0.20$, $K{=}20$, layers $\{8,16,20,24\}$.
IU-Xray is a public dataset, so images and reports can be reproduced here.
Cases span false cardiomegaly on a normal study, device recovery, borderline cardiomegaly with granuloma, and a moderate regression on an enlarged-heart study.
\textcolor{red}{Red}\,+\,bracket tags mark significant errors:
\textsf{\textbf{FF}} false finding,
\textsf{\textbf{MF}} missing finding.
\textcolor{teal}{Teal}\,+\,\textsf{\textbf{OK}} marks findings that match the reference.
\textcolor{orange}{Orange} marks residual significant issues in the steered output.
Gray \textsf{\textbf{INSIG}} / \textsf{\textbf{WS}} mark clinically insignificant wording differences (not counted toward GREEN).
All tags are derived from GREEN model raw completions.
\textbf{Improvement} (iuxray\_002410): steering removes a false cardiomegaly report, reaching GREEN\,=\,1.00.
\textbf{Improvement} (iuxray\_000286): the baseline omits the hemodialysis catheter; steering recovers the device and tip position (GREEN\,=\,0.80).
\textbf{Improvement} (iuxray\_001563): baseline falsely calls the heart ``within normal limits'' on a borderline-enlarged study with an RLL granuloma; steering moves heart size to "top normal" but still omits the nodule (GREEN\,=\,0.75).
\textbf{Regression} (iuxray\_000779): on a study with a slightly large heart, steering retains cardiomegaly but hallucinates a vertebral compression deformity (GREEN\,=\,1.00$\to$0.50), illustrating the FF trade-off (Table~\ref{tab:iuxray_pererror}).}
\label{tab:iuxray_qual_examples}
\end{table*}

\end{document}